\documentclass[10pt,journal,compsoc]{IEEEtran}
\ifCLASSOPTIONcompsoc
  \usepackage[nocompress]{cite}
\else
  \usepackage{cite}
\fi
\usepackage{amsmath}
\usepackage{epsfig}
\usepackage{graphicx}
\usepackage{amsmath}
\usepackage{amssymb}
\usepackage{flexisym}
\usepackage{breqn}
\usepackage{subfigure}
\usepackage{hyperref}
\usepackage{soul}
\usepackage[linesnumbered,ruled]{algorithm2e}
\usepackage[table]{xcolor}
\usepackage{colortbl}
\definecolor{Gray}{gray}{0.85}
\usepackage{array,multirow,graphicx}
\usepackage{xfrac}
\usepackage{booktabs}
\usepackage{breqn}
\usepackage{algorithmic}
\usepackage{xspace}
\usepackage[flushleft]{threeparttable}

\def\argmin{\mathop{\rm argmin}}

\newcommand{\real}{\ensuremath{\mathbb{R}}}

\newcommand{\R}{{\mathbb R}}

\usepackage{mathtools, stmaryrd}
\usepackage{xparse} \DeclarePairedDelimiterX{\Iintv}[1]{\llbracket}{\rrbracket}{\iintvargs{#1}}
\newcommand{\etal}{\textit{et al.}\@\xspace}

\newcommand*{\eg}{e.g.\@\xspace}
\newcommand*{\ie}{i.e.\@\xspace}
\ifCLASSINFOpdf
\else
\fi

\begin{document}
%
\title{ResNet-LDDMM: Advancing the LDDMM Framework Using Deep Residual Networks}
%
%
%
%

\author{Boulbaba Ben Amor,~\IEEEmembership{Senior,~IEEE}, Sylvain Arguillère and Ling Shao,~\IEEEmembership{Fellow,~IEEE}
\IEEEcompsocitemizethanks{\IEEEcompsocthanksitem B. Ben Amor is with the Inception Institute of Artificial Intelligence (IIAI), Abu Dhabi, United Arab Emirates; S. Arguillère is with the Laboratoire CNRS Paul Painlevé, Université de Lille, France; L. Shao is with Terminus Group, China.\protect\\
E-mail: boulbaba.amor@inceptioniai.org}
\thanks{Manuscript received February, 2021; accepted May, 2022.}}

\IEEEtitleabstractindextext{%
\begin{abstract}
In deformable registration, the Riemannian framework -- Large Deformation Diffeomorphic Metric Mapping, or LDDMM for short -- has inspired numerous techniques for comparing, deforming, averaging and analyzing shapes or images. Grounded in flows of vector fields, akin to the equations of motion used in fluid dynamics, LDDMM algorithms solve the flow equation in the space of plausible deformations, i.e. diffeomorphisms. In this work, we make use of deep residual neural networks to solve the non-stationary ODE (flow equation) based on an Euler’s discretization scheme. The central idea is to represent time-dependent velocity fields as fully connected ReLU neural networks (building blocks) and derive optimal weights by minimizing a regularized loss function. Computing minimizing paths between deformations, thus between shapes, turns to find optimal network parameters by back-propagating over the intermediate building blocks. Geometrically, at each time step, our algorithm searches for an optimal partition of the space into multiple polytopes, and then computes optimal velocity vectors as affine transformations on each of these polytopes. As a result, different parts of the shape, even if they are close (such as two fingers of a hand), can be made to belong to different polytopes, and therefore be moved in different directions without costing too much energy. Importantly, we show how diffeomorphic transformations, or more precisely bilipshitz transformations, are predicted by our registration algorithm. We illustrate these ideas on diverse registration problems of 3D shapes under complex topology-preserving transformations. We thus provide essential foundations for more advanced shape variability analysis under a novel joint geometric-neural networks Riemannian-like framework, \ie ResNet-LDDMM.                    
\end{abstract}

}

\maketitle

\IEEEdisplaynontitleabstractindextext

%
\IEEEpeerreviewmaketitle


\IEEEraisesectionheading{\section{Introduction}\label{sec:introduction}}

\IEEEPARstart{T}{he} deformable registration problem involves finding a single coordinate system in which to pinpoint several different shapes. This allows, for example, for the statistical analysis of shape data that takes into account their geometric properties. Many applications of this principle can be found in computational anatomy (CA), in which the shapes are extracted from medical images (MRI, PET scans, etc.) \cite{Haber04numericalmethods}. Various methods have been used to register different shapes or anatomical organs (see for example \cite{beg2005computing} and \cite{vialard2012diffeomorphic}). The LDDMM (large deformation diffeomorphic metric mapping) approach is one of the most popular for estimating plausible transformations (\ie diffeomorphisms). Taking advantage of the group structure of the manifold of diffeomorphisms of $\real^3$, it also comes with a proper metric for comparing shapes based on a certain kinetic energy of the deformation \cite{dupuis1998variational}.\\



\textbf{Problem formulation.} In the present work, we will restrict our study to discrete, unparametrized surfaces and more generally point sets/clouds of $\real^3$, denoted by $q=(x_1,x_2,\dots,x_n)^T\in \real^{n\times 3}$, $i\ne j \Rightarrow x_i \ne x_j $ ($n$ is the number of vertices/points in the point cloud or the meshed surface). In the language of shape analysis, this means that we work on the so-called spaces of landmarks. Given $q_S$ (source/template shape) and $q_T$ (target/reference shape), two point sets representing the same physical object or anatomical organ (liver, kidney, femur, heart, hippocampus, brain cortex, or simply a hand) of the human body, the goal is to find a reasonable transformation $\phi:\real^3\rightarrow\real^3$ such that a transformed version of the template shape $\phi.q_{S}=(\phi(x_1),\dots,\phi(x_n))^T$ is similar to the reference $q_{T}$. If the transformation $\phi$ is \textit{diffeomorphic} (i.e. smooth with smooth inverse), then $(\phi,q)\mapsto \phi.q$ is the associated \textit{group action} on the space of landmarks \cite{younes2010shapes}. In our special case of point sets, this smooth action is given by $\phi.q=(\phi(x_1),\dots,\phi(x_n))$. That is, the $i$-th landmark of $\phi.q$ is the position of the $i$-th landmark of $q$ after being moved in $\real^3$ by $\phi$. 

A common way to model the deformable registration problem is to consider the minimization of an energy functional (Eq. (\ref{Eq:min})) over the set of plausible deformations $\phi$,

\begin{equation} 
\mathcal{J}(\phi;q_S,q_T) = \underbrace{\mathcal{D}(\phi.q_S,q_T)}_{\text{Data term}} + \underbrace{\mathcal{R}(\phi)}_{\text{Regularizer}} 
\label{Eq:min}
\end{equation}

\noindent
where $\mathcal{D}(.,.)$ is the \textit{data term} that measures the discrepancy between the deformed shape $\phi.q_S$ and the target shape $q_T$. The term $\mathcal{R}(.)$ plays the role of a \textit{regularizer} and thus controls the plausibility of the solution $\phi^*$. In several applications, particularly when analyzing the anatomical parts/organs of the body, it is desirable to have a deformation $\phi$ that preserves local and global topology, preventing the deformation from creating holes or folding when applied to the source shape. This is true for \textit{diffeomorphisms}, for example. It is also true for the slightly more general \textit{bilipshitz maps}, which are essentially diffeomorphisms as well; they are homeomorphisms $\Phi$ such that both $\Phi$ and $\Phi^{-1}$ have a bounded rate of change (in particular, they are differentiable almost everywhere, with invertible differential).
LDDMM computes diffeomorphic transformations through the integration of smooth, time-dependent \textit{velocity fields} $f:[0,1]\times \real^3 \to \real^3$ over time \cite{beg2005computing}. Accordingly, time-dependent transformations $\phi:[0,1]\times \real^3 \to \real^3$ are derived. They are governed by an Ordinary Differential Equation, known as the \textit{flow equation} \cite{dupuis1998variational} formulated as in Eq. (\ref{Eq:FlowEq}),

\begin{equation}
\begin{split}
\dot\phi(t,x) = f(t,\phi(t,x))\textrm{,}\quad \phi(0,x) = x \\ \textrm{ for all } x \in \real^3 \textrm{ and } t \in [0,1]\textrm{,}\quad
\end{split}
\label{Eq:FlowEq}
\end{equation}
where $\dot\phi=\frac{\partial \phi}{\partial t}$ denotes the partial derivative over the variable time $t$ and $\phi(0,x)$ is the initial state taken to be the identity of $x$. Under adequate assumptions on $f$ (globally Lipschitz in space for fixed $t$, with a time-dependent Lipshitz constant integrable in time, for example), $f\mapsto\phi^f$ is a well-defined mapping into the space of time-dependent (essentially) diffeomorphisms of $\real^3$ by the Cauchy-Lipshitz theorem. A smoother $f$ also yields a smoother $\phi^f$ (and, hence, a diffeomorphism), as seen in \cite{trouve1995infinite} and \cite{dupuis1998variational} (see also \cite{younes2010shapes}, Theorem 8.7). The goal is then to find an optimal trajectory $\phi(t,.)$ connecting $\phi(0,.)=I_{\real^3}$ (starting point) to the end point $\phi(1,.)$ by finding a minimizer $f^*:[0,1]\times \real^3 \to \real^3$ of the following updated version of Eq. (\ref{Eq:min}), 

\begin{equation}
f^*=\argmin_{f(t,.)\in \cal{A}}\frac{1}{2\sigma^2}\mathcal{D}(\phi^f(1).q_{S},q_{T})
+\frac{1}{2} \int_{0}^{1}  \| f(t,.) \|_{\cal{A}}^2 dt.
\label{Eq:LDDMM}
\end{equation}
Here, $\cal{A}$ is the space of \textit{admissible} velocity fields (i.e. with the smoothness condition) which give rise to $\textrm{Diff}_{\cal{A}}(\real^3)$, defined by
$
\textrm{Diff}_{\cal{A}}(\real^3) = \{\phi^f(1), f\in L^1([0,1],\cal{A})\}$, the Group of Diffeomorphisms associated to $\cal{A}$. The weighting factor $1/2\sigma^2$ in Eq. (\ref{Eq:LDDMM}) balances the influence of the data attachment term $\cal D$ and the regularizer term $\cal R$ of the general formulation in Eq. (\ref{Eq:min}). We denote by $\phi^f(1)$ the final diffeomorphism (at time $t=1$) to transform $q_{S}$ to have $\phi^f(1).q_{S} \sim q_{T}$. 

From the formulation above, we obtain the following interesting geometric interpretations and properties: (1) The quantity ${\cal S}_t=\| f(t,.) \|^2_{\cal{A}}$ is the \textit{kinetic energy} of the whole system at the time step $t$ (refer to \cite{younes2010shapes}); (2) For $q,\tilde q$ two sets of landmarks with the $n$ points, $n\in \mathbb{N}$,
$$
d_{\cal{A}}^n(q,\tilde q)= \inf_{f(t,.)\in \cal{A}} \{\int_{0}^{1}  \| f(t,.) \|_{\cal{A}} dt, \tilde q= \phi^f.q\}
$$
is a \textit{metric} on the space of landmarks with $n$ points. One can even deduce a metric $d_{{\cal A}}$ on $\textrm{Diff}_{\cal{A}}$ itself in a similar way, making $(\textrm{Diff}_{\cal{A}},d_{{\cal A}})$ a \textit{complete metric space} (\cite{trouve1995infinite} and \cite{younes2010shapes}, Theorem 8.15). This is a very important characteristic of the LDDMM family of methods and can be used to assess the similarity of various objects by analysis of the velocity fields $f$ that induce the transformation $\phi^f(1)$, which aligns them.


This elegant LDDMM framework has served as a starting point for several approaches in the literature which can be categorized into: \textit{(i) relaxation methods} (\eg \cite{beg2005computing}), which compute velocities for multiple points in time, and \textit{(ii) shooting methods} (\eg \cite{vialard2012diffeomorphic}) which take advantage of the conservation of momentum (and in particular a constant kinetic energy ${\cal S}^*_t=\| f^{*}(t,.) \|^2_{\cal{A}}$) and determine the evolution of the transformed template based solely on the initial velocity.
Importantly, both approaches define an admissible Hilbert space $\cal A$ of velocity fields as an Reproducing Kernel Hilbert Space (RKHS). The kernel is taken to be the Green’s kernel $K=L^{-1}$, (where $L$ is a differential operator which defines the inner product of $\cal A$ as well as the induced norm $\|.\|_{\cal A}$) \cite{younes2010shapes}. In practice, smoothing of the velocity fields is achieved by convolution with suitable kernels, \eg, Gaussian kernels with positive scale. With this definition of $\cal A$ as an RKHS, we complete the formulation of the LDDMM framework.      


\section{Related Work}
\label{Sec:RelatedWork}

In this section, we first review different variants of the LDDMM algorithm. Then, we discuss recent techniques developed for \textit{non-rigid point clouds} registration (without any point connectivity prior). Third, we focus on the \textit{Functional Maps} framework and its \textit{Deep Learning} variants. Finally, we discuss the connection between LDDMM for shape registration, Residual Networks and geometric flows on which our joint ResNet-LDDMM framework grounds.

\subsection{Large Deformation Diffeomorphic Metric Mapping}
\label{sec:LDDMM}

Under the LDDMM formulation, one can distinguish two paradigms \cite{polzin2020discretize}. Firstly, the \textit{optimize-then-discretize} schema which first derives the \textit{Hamiltonian} of the continuous problem, then, deduces optimality conditions of continuous optimization problems using the calculus of variations (\eg \cite{arguillere2016multipleshapes}, \cite{arguillere2016diffeomorphic} \cite{younes2009evolutions}, \cite{gris2018sub}, \cite{vialard2012diffeomorphic}, \cite{charon2013varifold}, \cite{miller2015hamiltonian}, and \cite{miller2020coarse}). Ultimately, these conditions are discretized and solved, typically using Gradient Descent. LDDMM algorithms (both \textit{relaxation} and \textit{shooting} methods) achieve accurate registration results, but are costly in running time (counting hours on CPU) and memory. As a response to this, the \textit{discretize-then-optimize} paradigm was born. It involves first discretizing the objective functional and constraining equations, and then solving the discrete optimization problem using numerical optimization methods (\eg \cite{modersitzki2009fair}, \cite{mang2017lagrangian} and \cite{polzin2020discretize}).

Similar to LDDMM, the SVF (Stationary Velocity Fields) framework is an alternative for finding a diffeomorphic transformation between shapes. It was first introduced in \cite{arsigny2006log}. As LDDMM, SVF works on a vector space of images and a Lie group of diffeomorphic transformations. SVF generalizes the principal logarithm to non-linear geometrical deformations which falls into parameterizing diffeomorphisms with stationary speed vector fields. Despite its satisfactory results in practice, the logarithm is well-defined only for transformations close enough to the identity (as the Lie group exponential map is usually not surjective), which makes the registration results under large deformations uncertain. That is, the optimal transformation is not smooth with regard to the images, so that a small change in images may lead to a large change in the path connecting them \cite{kobatake2017computational}. Furthermore, the underlying space (of the SVF framework) is not a Riemannian manifold and there is no Riemannian metric, geodesic, or connection involved \cite{kobatake2017computational}. While an LDDMM curve is obtained by integrating a time-dependent vector field specified by the Riemannian metric, an SVF curve is an integral curve of a stationary vector field, in the corresponding Lie algebra. Hence, SVF works on the structure of the Lie group of diffeomorphic transformations instead of the underlying Riemannian manifold. 
 

\subsection{Non-rigid \textit{Point Clouds} Registration Methods}
\label{sec:PointCloud}
Less constrained, but more efficient approaches have  also been developed with the increasing amount of 3D data available and depth-sensors. These approaches solve the problem of matching two 3D point clouds in the presence of non-rigid deformations. They typically adopt the $\ell_p$ type robust estimator to regularize the fitting and smoothness. For instance, Amberg \etal~ (who developed an $\ell_2$-regularization method) have extended in \cite{amberg2007optimal} the popular ICP (iterative closest point) algorithm to cover non-rigid transformations. They include in their framework different regularizations, as long as they have an adjustable stiffness parameter. So, thus, the registration loops over a series of decreasing stiffness weights, and incrementally deforms the template towards the target, recovering the whole range of global and local deformations. However, the presence of noise, outliers, holes, or articulated motions between the point clouds can potentially result in alignment errors (we will therefore refer to this approach as \textit{N-ICP}). Taking another direction, Li \etal proposed in \cite{li2018robust} the \textit{RPTS} method with $\ell_1$-regularization using re-weighted sparsities on positions and transformations to estimate the deformations on point clouds. They formulated the energy function with dual sparsities on both the data term and the smoothness term, and defined the smoothness constraint using local rigidity. Starting from the observation of the existence of an intrinsic articulated subspace in most non-rigid motions, Guo \etal \cite{guo2015robust} proposed an $\ell_0$-based motion regularizer with an iterative optimization solver that can implicitly constrain local deformation only on joints with articulated motions. As a consequence, the solution space is reduced to physical plausible deformations. We will refer to this approach as \textit{SVR-}$\ell_0$. Recently, in \cite{Yao_2020_CVPR}, Yao \etal proposed a formulation based on a globally smooth robust estimator for data fitting and regularization, which can handle outliers and partial overlaps by enforcing sparsity using the \textit{Welsch’s function}. They made use of the majorization-minimization (MM) algorithm to tackle the problem, which reduces each iteration to solve a simple least-squares problem with L-BFGS. Their approach (which we will call \textit{QNS}) achieves lower registration errors compared to the previous approaches. While these approaches bring efficient solutions to solve the correspondence problem, they often require good initialization, i.e., a set of labeled landmarks on both meshes. Further, they do not guarantee diffeomorphic transformations between 3D shapes and registering a shape A to a shape B, and inversely, can yield in different matching results.

Following the \textit{N-ICP} approach \cite{amberg2007optimal}, Dyke \etal~ extend the non-rigid registration to better account for anisotropic deformations (e.g., stretches) \cite{dyke2019non}. In order to address (local) anisotropic deformations, the method iteratively estimates local anisotropy (represented as principal directions and principal scaling factors), which is then incorporated in an extended diffusion pruning framework to identify consistent correspondences, taking anisotropy into account when calculating geodesic distances. Taking a different direction, Groueix \etal~ \cite{groueix20183d} train a point cloud Auto-Encoder, termed 3D-CODED, to compute features in order to match a template to input shapes. This approach required a huge amount of training data.

\begin{figure*}[ht!]
  \centering
  \includegraphics[width=\linewidth]{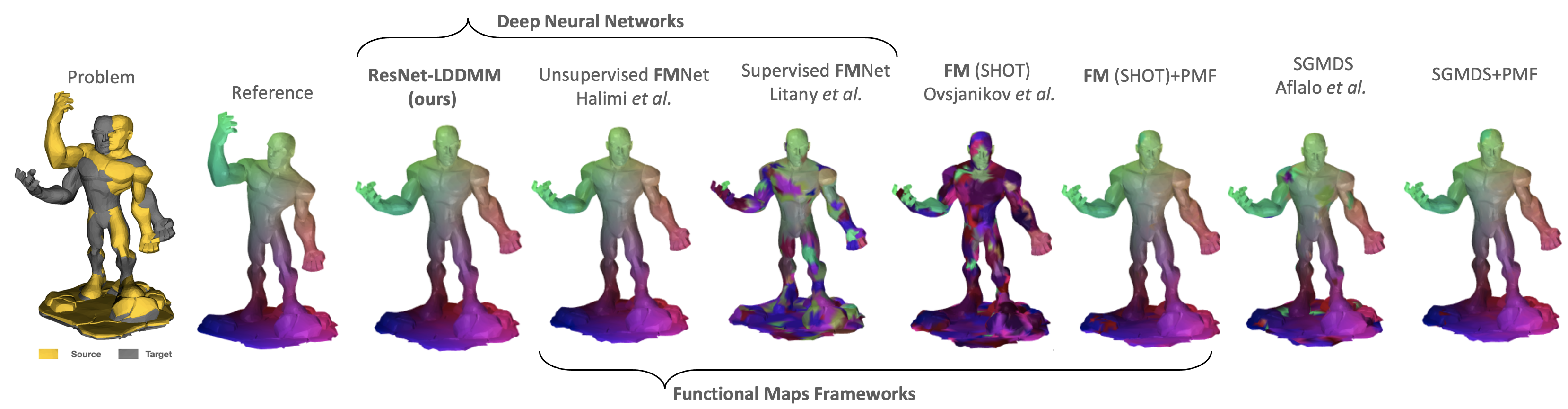}
\vspace{-0.8cm}
  \caption{Qualitative comparison of ResNet-LDDMM (third column) to state-of-the-art approaches including \textit{Functional Maps (FM)} framework and Supervised and Unsupervised \textit{FM}Net on human-like artistic models (the six last registration results are previously reported by Halimi \etal in \cite{halimi2019unsupervised}).}
   \label{Fig:Artist}
\end{figure*}

\subsection{The \textit{Functional Maps} Framework}
\label{sec:FunctionalMaps}

Rather than matching individual points, which remains a hard problem, a family of approaches propose to match scalar functions over the shapes. Among them, the \textit{Functional Maps} framework introduced by Ovsjanikov \etal in \cite{ovsjanikov2012functional}. The key idea is to find shape-to-shape correspondence using real-valued (descriptor) functions. Candidate function maps are ideally invariant under the transformations to filter out. Basically, a shape map lifts to a linear operator mapping the function spaces. Then, a compact representations of functional maps are obtained by projection of the linear operator in a predefined basis (the \textit{Laplace-Beltrami} eigenbasis (LBO) is a typical choice). The fundamental problem is to obtain such suitable descriptors.

FMNet introduced in \cite{litany2017deep} learns descriptors to achieve the correspondence. So that, the output of FMNet is a dense vector-valued descriptor, calculated on each of the input shapes. Recently, in \cite{halimi2019unsupervised} the first unsupervised learning scheme was proposed. Dense correspondence is obtained by minimizing pair-wise geodesic distance distortion computed on the pair of shapes. This approach could be viewed as the fusion of GMDS \cite{aflalo2016spectral} and an unsupervised version of FMNet \cite{litany2017deep}. The Deep Learning Functional Maps approaches require an initialization with pre-computed descriptors typically SHOT \cite{tombari2010unique}. In \cite{halimi2019unsupervised}, the descriptors learned from correspondences, are replaced by geometric invariant descriptors (matrices of all pairwise geodesic distances). More recently, inspired by FMNet (unsupervised regime) and 3D-CODED (supervised regime), a deep geometric functional map (\ie an hybrid approach) was proposed in \cite{donati2020deep}. Here, the network performs a feature-extraction and learns directly from raw shape geometry, combined with a novel regularized map extraction layer and loss, based on the functional map representation. This hybrid approach is less dependant to the data size in the training set compared to the 3D-CODED approach \cite{groueix20183d}. Other approaches first map the shapes to a canonical space (\eg \cite{ovsjanikov2010one}), then solve for the correspondence between obtained parametric representations, blend across such maps \cite{kim2011blended}, or compute multiple near-isometric dense correspondences \cite{ren2020maptree}, rather than optimizing for a single solution.

\subsection{Link between ResNets and Geometric Flows}
\label{sec:LinkResNet-LDDMM}

The interpretation of ResNets as transformations on the data space is not new. For example, \cite{NIPS2017_0ebcc77d} studies the induced change on the Riemannian metric of the deformations, while \cite{ruthotto2019deep} interprets convolution of images as partial differential operators, and each residual layer as a forward Euler scheme. From there, the recent paper \cite{rousseau2020residual} of Rousseau \etal uses for the first time the idea that ResNets actually generate diffeomorphisms on the space in which the data lives (or at least, bi-lipshitz maps). The paper both deepens the theoretical understanding of ResNets and provides improved algorithms for the classification of data. In the same context, L. Younes have introduced a diffeomorphic learning paradigm in which training data is transformed by a diffeomorphic transformation (defined in high dimensional space) before the final prediction \cite{younes2019diffeomorphic}. Importantly, an Optimal Control formulation of the learning algorithms was derived as well as the importance of such smooth and invertible transformations in generative models was discussed. 

The main difference between our approach and that of  \cite{rousseau2020residual}, which also defines diffeomorphisms, is that we do not build diffeomorphisms on the data space (that is, the space $\real^{n\times 3}$ in which the whole shape lives), but on the ambient space $\real^3$. Then, this transformation is applied separately to each point $x_i, i=1,\dots,n$ of the shape. This ensures the preservation of the topology of the shape. This is partly why we did not include surface convolution layers with filter size $>1$, for example. Such layers would still yield diffeomorphisms in  $\real^{n\times 3}$ as per \cite{rousseau2020residual}, but those transformations would not translate into well-defined transformation of $\real^3$.


\subsection{Contributions and Paper Organization}

In this work, we propose a novel joint geometric-neural networks framework for diffeomorphic registration of 3D point clouds (and meshed surfaces). Our formulation differs in several aspects from previous deep learning approaches. It is also much more in line with LDDMM methods than \cite{rousseau2020residual}. The paper's contributions are summarized as follows,

$\hookrightarrow$ We introduce a Riemannian-like framework (thus, a length space), for deforming, registering and comparing 3D shapes, by revisiting the original LDDMM framework using Deep Residual Networks. Up to our knowledge, despite the similarity to previous interpretations of Residual Networks as flow of diffeomorphisms, our work is the first to demonstrate these ideas on 3D shape registration problems.
    
$\hookrightarrow$ Our ResNet-LDDMM predicts time-dependent, regular and smooth velocity fields (the endpoint of the flow is the desired diffeomorphic transformation). It brings a novel regularization driven by the time-integrated kinetic energy (or the geometric norm of the control, \ie the velocity fields). We demonstrate that just like in LDDMM, ResNet-LDDMM does generate (essentially) a diffeomorphism of the ambient space $\real^3$ itself, instead of just on the data space as in \cite{rousseau2020residual}.
    
$\hookrightarrow$ Appropriately applying the ReLU activation function to predict velocity fields, we reveal how each time-step of our ResNet-LDDMM divides the space $\real^3$ into multiple polytopes (\ie, unbounded polyhedra) in which optimal affine velocity fields are computed. This ensure more flexibility in the registration while maintaining diffeomorphic properties. Up to our knowledge, this geometric interpretation has not been described so far, although some similar interpretations have been suggested in \cite{rousseau2020residual} and \cite{younes2019diffeomorphic} in the context of diffeomorphic learning. 

$\hookrightarrow$ We report competitive results in comparison to the original LDDMM framework, to recent non-rigid point-clouds registration approaches, to different variants of the \textit{Functional Maps} framework and to supervised Deep Learning techniques. As a first experimental illustration of our non-rigid registration approach, we report in Fig.\ref{Fig:Artist} results on humanoid-like meshes and compare with state-of-the-art methods. This figure highlights the limitation of supervised learning techniques \cite{litany2017deep} to generalize on new deformations, as discussed in \cite{halimi2019unsupervised}. In addition, as experimented in \cite{halimi2019unsupervised}, Functional Maps require, in general, a time-consuming post-processing refinement step \cite{vestner2017efficient}, called Product Manifold Filter (PMF), and an initialization step using SHOT descriptors. More efficient and accurate refinement methods have been developed recently. For example, in \cite{ezuz2017deblurring} a new method for map deblurring was proposed, by introducing a smoothness prior on the reconstructed map which results in better maps, especially when the shapes are not isometric. Taking a different direction, Melzi \textit{et al.} showed in \cite{melzi2019zoomout} that a higher resolution map can be recovered from a lower resolution one through a remarkably simple and efficient iterative spectral up-sampling technique. Consequently, a simple and efficient method for refining maps or correspondences by iterative upsampling in the spectral domain have been proposed. More recently, Pai \textit{et al.} introduced in \cite{pai2021fast} a method which combines the simple and concise representation of correspondence using FMs with the matrix scaling schemes from computational optimal transport. The approach improves accuracy and bijectivity of correspondences.

While our ResNet-LDDMM operates on the original point clouds, \textit{Functional Maps} approaches require to define (\eg \cite{halimi2019unsupervised}) or to lean (\eg \cite{litany2017deep}) a descriptor function, which make them conceptually different. The rest of the paper is organized as follows. In Sec. \ref{sec:ResNet-LDDMM}, we describe our ResNet-LDDMM framework from both geometric and (unsupervised learning) neural networks perspectives. How ResNet-LDDMM naturally builds diffeomorphic transformations between shapes is formally elaborated in Sec. \ref{sec:Diffeos}. Furthermore, we provide a geometric description of the velocity fields accompanied by a qualitative interpretation on the registration methodology behind ResNet-LDDMM. Sec. \ref{sec:Experiments} is dedicated to multiple experiments. Some concluding remarks and perspectives are drawn in Sec. \ref{sec:Conclusion}.



\begin{figure*}
  \centering
   \includegraphics[width=\linewidth]{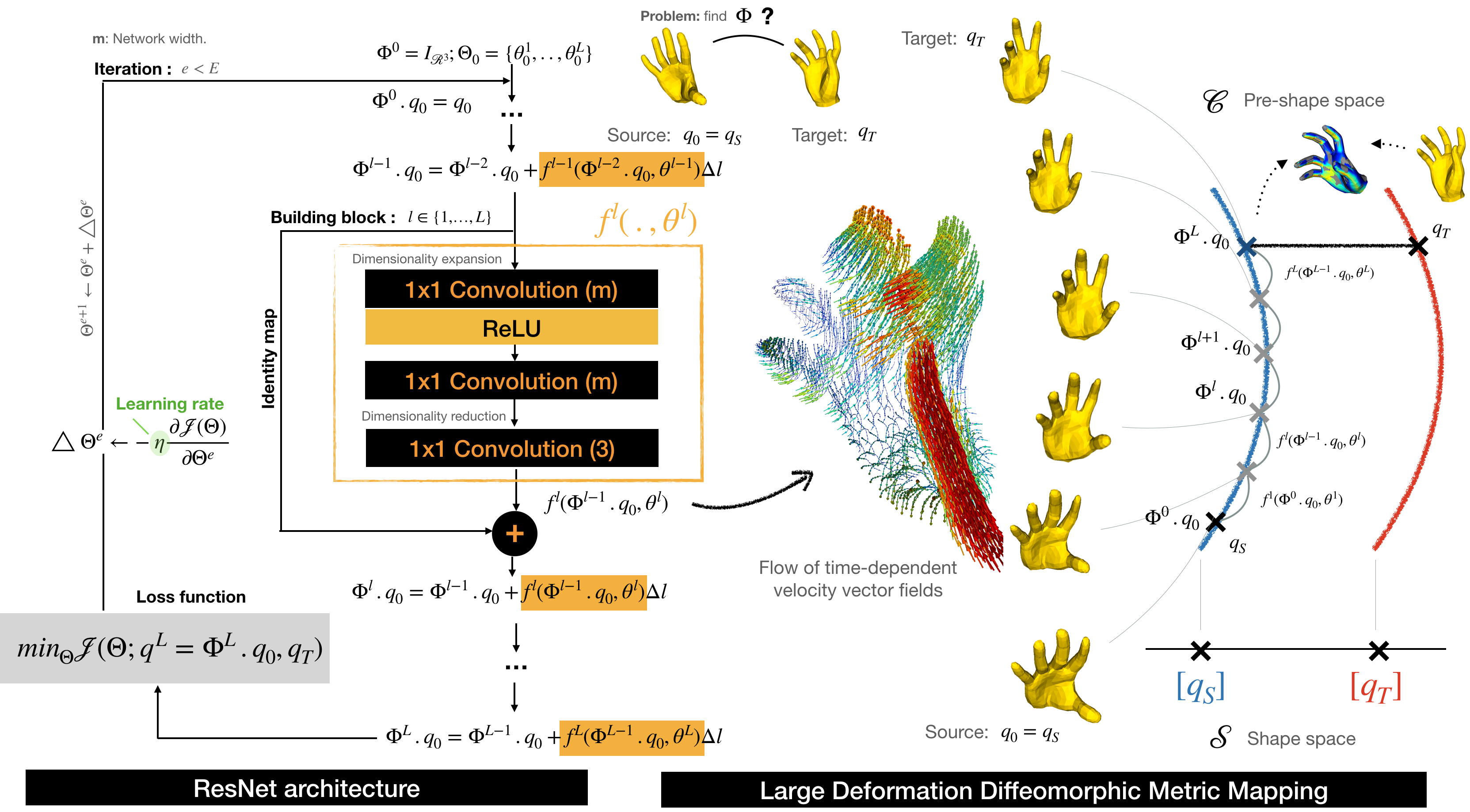}
  \caption{Overview of our joint ResNet-LDDMM diffeomorphic registration framework -- the ResNet architecture is shown on the left, comprising of an ensemble of successive two-layers ReLU networks followed by a dimensionality reduction layer (the ensemble is called a building block). Each predicts a time-dependent velocity field $f^l(.,\theta^l)$. On the right, we show the flow of obtaining velocity fields and diffeomorphisms $\Phi^l$ applied to the source shape, including the end-point $\Phi^L$, achieved by integration. In Computational Anatomy (\eg \cite{miller2004computational}), it is assumed that observations of the same anatomical organ are elements of a common shape Orbit, so that $[q_S]$ and $[q_T]$ coincide. The Network's width $m$ is the number of filters.}
   \label{Fig:Overview}
\end{figure*}
\section{ResNet-LDDMM, or LDDMM revisited}
\label{sec:ResNet-LDDMM}

The central idea in our ResNet-LDDMM is to make use of a particular family of deep neural networks, i.e. residual deep networks (or ResNets), to solve the flow equation Eq. (\ref{Eq:LDDMM}). In supervised learning, a residual network \cite{DBLP:conf/cvpr/HeZRS16,DBLP:conf/eccv/HeZRS16} has the following form:

\begin{equation}
h^{l+1} = h^{l} + f^{l}(h^{l},\theta^{l})    
\label{Eq:hidden}
\end{equation}

\noindent where $h^{l}$ is the trainable hidden layer, $l\in \{1,\dots,L\}$, and $\theta^l\in \{\theta^1,\dots,\theta^L\}$ denotes the network parameters. Compared to a standard feed-forward network, where $h^{l+1} = f(h^{l},\theta^{l})$, at the heart of ResNet is the ultimate idea that every additional layer should contain the identity function as one of its elements. Thus, a residual $h^{l+1}-h^{l}$ is learned instead of transforming the output of the previous layer $h^{l}$ to $h^{l+1}$. Recent works have pointed out a striking similarity between this important property in residual networks and the numerical solution of ordinary differential equations (e.g. \cite{weinan2017proposal,haber2017stable,ruthotto2020machine}) using the forward Euler method with a given initial value. In fact, training a deep residual network is viewed as a discretization of a dynamical training system governed by a first-order ODE, where the network layers are viewed as time-steps and the network parameters $\theta^l$ are interpreted as the \textit{control} to optimize \cite{liu2019deep}. 

\begin{figure*}[ht]
  \centering
   \includegraphics[width=\linewidth]{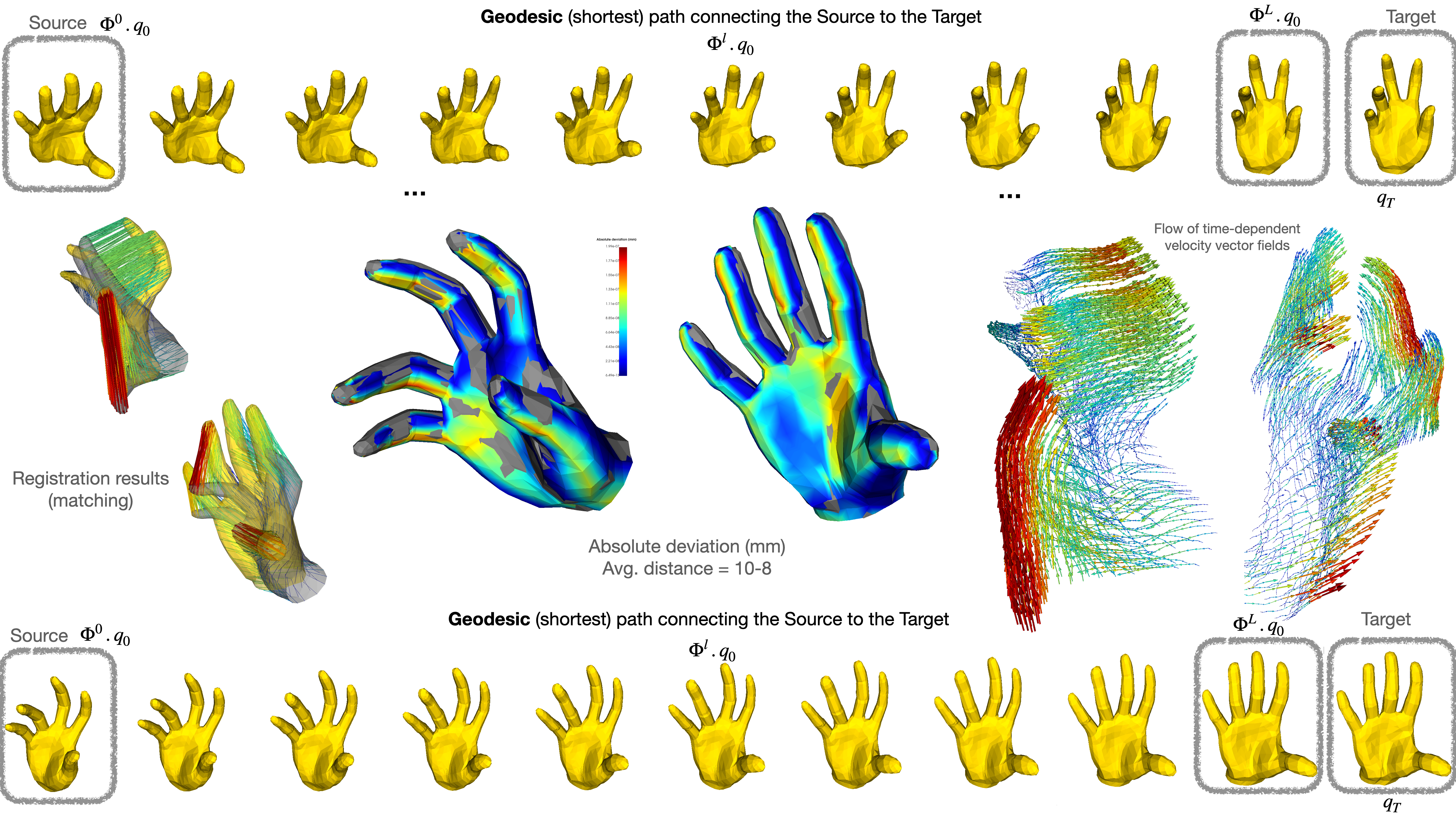}
  \vspace{-0.5cm}
  \caption{A pictorial summary of our diffeomorphic registration results. Top: a geodesic path connecting a source hand shape to a target shape; Bottom: the inverse path (source and target change roles); Center-right: both flows of the time-dependent velocity fields $f^l(.,\theta^l)$, building blocks of our ResNet-LDDMM. Center-left: final Euclidean displacements and registration results.}
   \label{Fig:Example}
\end{figure*}

Inspired by the general formulation and aiming to solve the flow equation Eq. (\ref{Eq:FlowEq}), we build our ResNet-LDDMM framework. As shown in Fig. \ref{Fig:Overview}, our ResNet-LDDMM is an ensemble of $L$ successive identical building blocks. Each building block is composed of three fully connected layers and a point-wise ReLU activation function separating the first two layers. From the LDDMM perspective (right panel of Fig. \ref{Fig:Overview}), each building block ${f^l}$ represents a velocity vector field $\real^3\rightarrow\real^3$ at a discrete time $l$. The $f^l$ are   parameterized by $\theta^l$. If $f^l$ are sufficiently smooth and spatially regular (we postpone this demonstration to Sec. \ref{sec:Diffeos}), one can then build a diffeomorphic transformation $\Phi^f(1)$ (by integration of Eq. (\ref{Eq:ResNet})) that moves the initial shape $q_S$ to fit $q_T$, which is mathematically equivalent to $\Phi(1).q_S \sim q_T$ and  $\Phi(1)=\Phi^L$:

\begin{equation}
\Phi^{l+1}.q_0 = \Phi^{l}.q_0 + \Delta^L f^l(\Phi^{l}.q_0,\theta^{l}),\ \text{with}\ \Delta^L=\frac{1}{L}.
\label{Eq:ResNet}
\end{equation}

Compared to the conventional LDDMM formulation, here, the time-varying velocity vector fields are deep neural networks $f^l(.,\theta^l)$ that should be optimized with respect to a fit error. To this end and if assume a ResNet-LDDMM with number of blocks $L \to \infty$ (i.e. $\frac{1}{L}\to 0$), we cast Eq. (\ref{Eq:LDDMM}) as,  
\begin{equation}
\begin{split}
\Theta^*, \Phi^*(1)=\argmin_{\Theta=\{\theta^t\}_{t \in [0,1]}}\frac{1}{2\sigma^2}\mathcal{D}(\Phi^{\Theta}(1).q_S,q_{T})\\
+\frac{1}{2} \int_{0}^{1}  \| f^t(\Phi^{\Theta}(t).q_S,\theta^{t}) \|_{\ell^2}^2 dt, 
\end{split}
\label{Eq:ResNet-LDDMM}
\end{equation}

\noindent subject to $\partial_t \Phi^{\Theta}(t,.)=f^t(\Phi^{\Theta}(t,.),\theta^{t})$ for a.e. $t \in [0, 1]$ and $\Phi(0,.)$ denotes the identity map (the neutral element of the group $\textrm{Diff}(\real^3)$). The family $f^t(.,\theta^t)$ is the ensemble of functions to be approximated by the network which coincides with the time-dependent velocity fields along the time interval $[0,1]$ and $\Theta=(\theta^t)_{t \in [0,1]}$ are the whole ResNet-LDDMM parameters. $\Phi^{\Theta}(1)$ is the final transformation and $\Phi^{\Theta}(t)$ are the intermediate transformations. The choice of ${\cal D}(.,.)$ will be detailed in the next paragraph. The second term is a regularizer that presents the length of the path connecting $q_S$ to $\Phi^{\Theta}(t).q_S$, on the orbit of $q_S$, which ensures that the optimal vector field stays regular enough to ensure topology preserving transformations (Sec. \ref{sec:GeometricRegularization}). In the right panel of Fig. \ref{Fig:Overview} we illustrate how ResNet-LDDMM enables a minimizing path (a geodesic) on the orbit of $q_s$. Intermediate shapes are discrete elements of the geodesic outputs of the action of the intermediate diffeomorphisms $\Phi^l$ on the source shape $q_S$. $\Phi^l$ are deduced from the optimal vector fields $f^l(.,\theta^l)$ sitting in between the panels of Fig. \ref{Fig:Overview} (colors (cold $\to$ hot) reflect the absolute amount of displacement at each time-step and at a vertex-level).


\textbf{ResNet-LDDMM Network Architecture.} Let us now come back to the architecture of our ResNet-LDDMM. As illustrated in the left panel of Fig. \ref{Fig:Overview}, our network takes as inputs (1) the source shape $q_S$, or to be more accurate $\Phi^{0}.q_S$, where $\Phi^{0}$ is the identity diffeomorphism of $\real^3$, and (2) an initial guess $\Theta_0=(\theta_0^1,\dots,\theta_0^l)$, initial parameters of the network and (3) the target shape $q_T$ to evaluate the loss function Eq. (\ref{Eq:ResNet-LDDMM}). The \textit{velocity fields} $f^{l}(\cdot,\theta^l)$ and the corresponding diffeomorphisms $\Phi^{l}$ are predicted by the $l$-indexed successive building blocks of the network (orange box in Fig. \ref{Fig:Overview}). The number of building blocks $L$ in the ResNet-LDDMM coincides with the total number of time-steps in the conventional LDDMM, once discretized. 

-- Each \textbf{building block} (denoted by $f^l(.,\theta^l)$ and generating the $l$-th velocity field on $\real^3$) consists of three successive fully connected weight layers. Weight matrices $W_i$ and bias vectors $b_i$ (limited to the first and second weight layers) define point-wise convolution operations, with kernel filters of size $1\times 1(\times m)$, performed separately on all points. This is equivalent to designing a fully connected layer as described in the Network-in-Network \cite{lin2014network}. We will denote $m$ as the width of individual building blocks. Subsequently, a ReLU activation function is applied to the output of the first weight layer within each building block. The last weight layer reduces the dimensionality of the output of the previous layer to get velocity vectors in $\real^3$ (so, the width of this layer is exactly $3$).  While the weight layers conduct affine transformations of the input (output of the previous building block), ReLU reduces to zeros all negative outputs of the first layer, thus dividing the space $\real^3$ into $m$-half spaces for more flexible prediction of the velocity fields (this point will be detailed in Sec. \ref{sec:GeometricDescription}). 

-- The \textbf{identity map}, which connects successive building blocks of our ResNet-LDDMM allows the network to focus predicting the difference $\Phi^l-\Phi^{l-1}$, and thus time-dependent velocity fields. This specific architecture allows a forward Euler scheme to solve the non-stationary ODE (flow equation Eq. (\ref{Eq:FlowEq})). Intuitively, an individual building block $f^l$ will move each point $x^l\in q^l$ into $x^{l+1}\in q^{l+1}$ such that passing the three successive connected layers results in

\begin{equation}
\begin{split}
x^{l+1}-x^l=&f^l(x^l,\theta^l).\Delta^L
\\=& W^l_3(W^l_2(ReLU(W^l_1x^l+b^l_1)+b^l_2).\Delta^L
\end{split}
\end{equation}

\noindent Here $W^l_i$ are the set of filters and $b^l_i$ biases which compose the weight layers.

-- The \textbf{loss function} ${\cal J}(\Theta$) given in Eq. (\ref{Eq:ResNet-LDDMM}) to be minimized consists of a data term and the integration of instantaneous kinetic energy terms as a regularization term. While the data term pushes the deformed template $\Phi^{L}.q_S$ closer to the target $q_T$ by minimizing an appropriate distance between point clouds, the regularization controls activities of the building blocks, i.e. their outputs. These terms are balanced with the $1/2\sigma^2$ parameter, as defined in Eq. (\ref{Eq:ResNet-LDDMM}).

-- As a \textbf{data attachment} term, we use either a point-wise distance called the \textit{Chamfer's distance (CD)} (Eq. (\ref{Eq:Chamfer})) or a global distance measure referred to us the \textit{Earth mover’s distance (EMD)}, the \textit{Wasserstein distance} or the \textit{Optimal Transport (OT) costs} (Eq. (\ref{Eq:Wasserstein})). The former measures the squared distance between each point in one set to its nearest neighbor in the other set and vice verse. Mathematically, it is formulated as follows: Eq. (\ref{Eq:Chamfer}),

\begin{equation}
    {\cal D^{CD}}(q_1,q_2)=\sum_{x\in q_1} \min_{y \in q_2} \|x-y\|^2_2 + \sum_{x\in q_2}\min_{y \in q_1} \|x-y\|^2_2.
\label{Eq:Chamfer}    
\end{equation}

The second distance first introduced in \cite{rubner2000earth}, measures the discrepancy between distributions when accounting for their respective geometries \cite{feydy2019interpolating}, defined by Eq. (\ref{Eq:Wasserstein}):

\begin{equation}
{\cal D^{EMD}}(q_1,q_2)=\min_{\xi}\sum_{(x,y)\in q_1\times q_2} \|x-y\|^p_2\xi(x,y).
\label{Eq:Wasserstein}    
\end{equation}

\noindent The \textit{Earth Mover's distance} ${\cal D^{EMD}}$ corresponds to $p=1$ and is based on finding a transport plan $\xi: q_1\times q_2\rightarrow [0,1]$, such that $\sum_{x\in q_1} \xi(x,y^*)=\frac{1}{n_2}$ and $\sum_{y\in q_2} \xi(x^*,y)=\frac{1}{n_1}$ for all $x^*$ in $q_1$ and $y^*$ in $q_2$, with $n_1$ and $n_2$ respectively the number of points in $q_1$ and $q_2$. The function $\xi(x,y)$ represents the proportion of mass at $x$ that is transported to $y$. One can also take $p=2$ instead, which corresponds to the \textit{Wasserstein} distance.
In practice, the iterative \textit{Sinkhorn’s Algorithm} is used to approximate the EMD distance, i.e. an Optimal Transport with Entropic Constraints (the reader is directed to \cite{NIPS2013_af21d0c9} and \cite{chizat2020faster} for more details). Importantly, we note that both distances are differentiable almost everywhere \cite{feydy2017optimal}. The choice of distance used in the data term depends on the data itself and the nature of the deformations, as we will discuss in Sec. \ref{sec:Experiments}.   
\begin{figure}[ht]
  \centering
  \includegraphics[width=\linewidth]{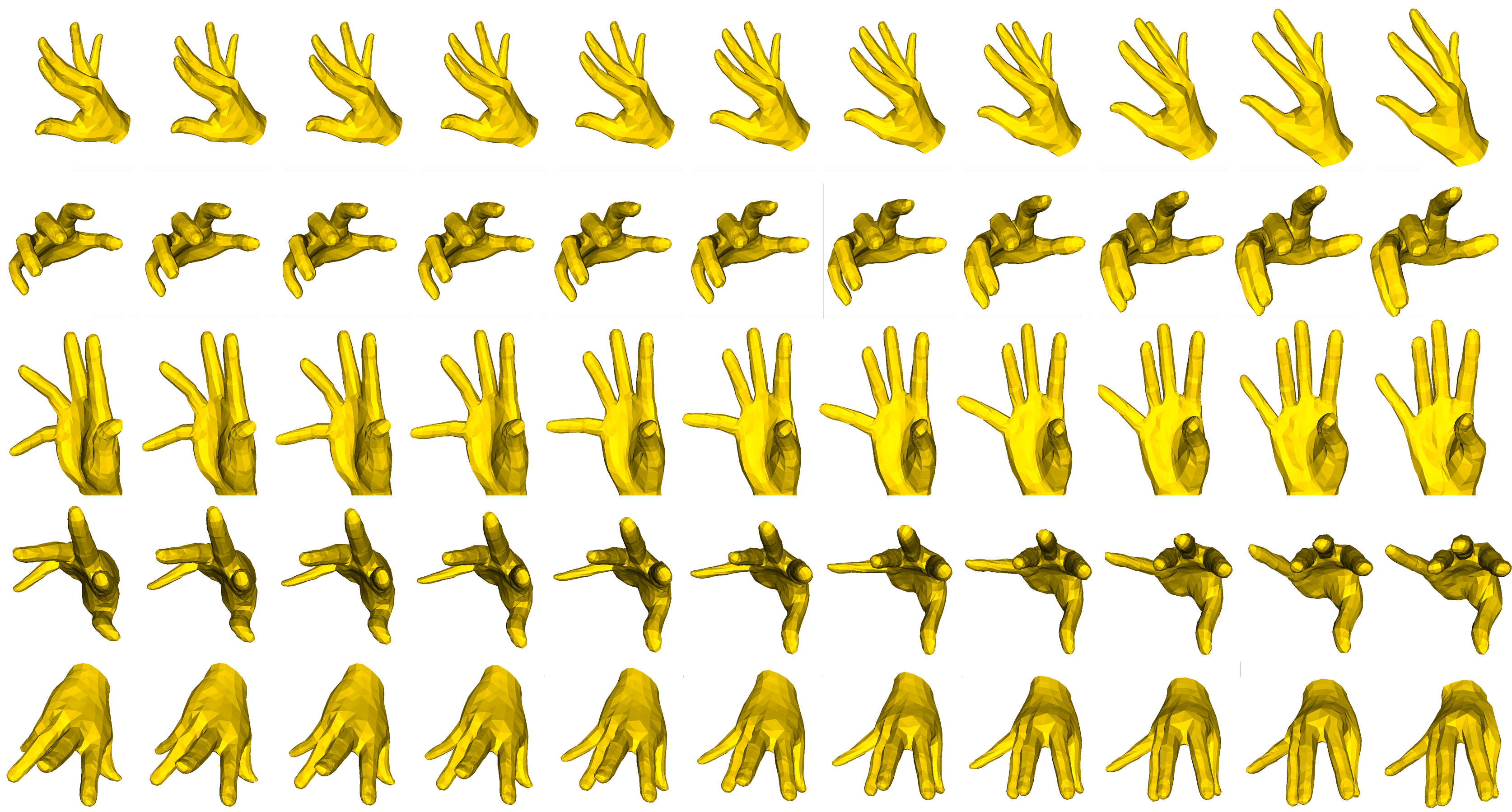}
   \includegraphics[width=\linewidth]{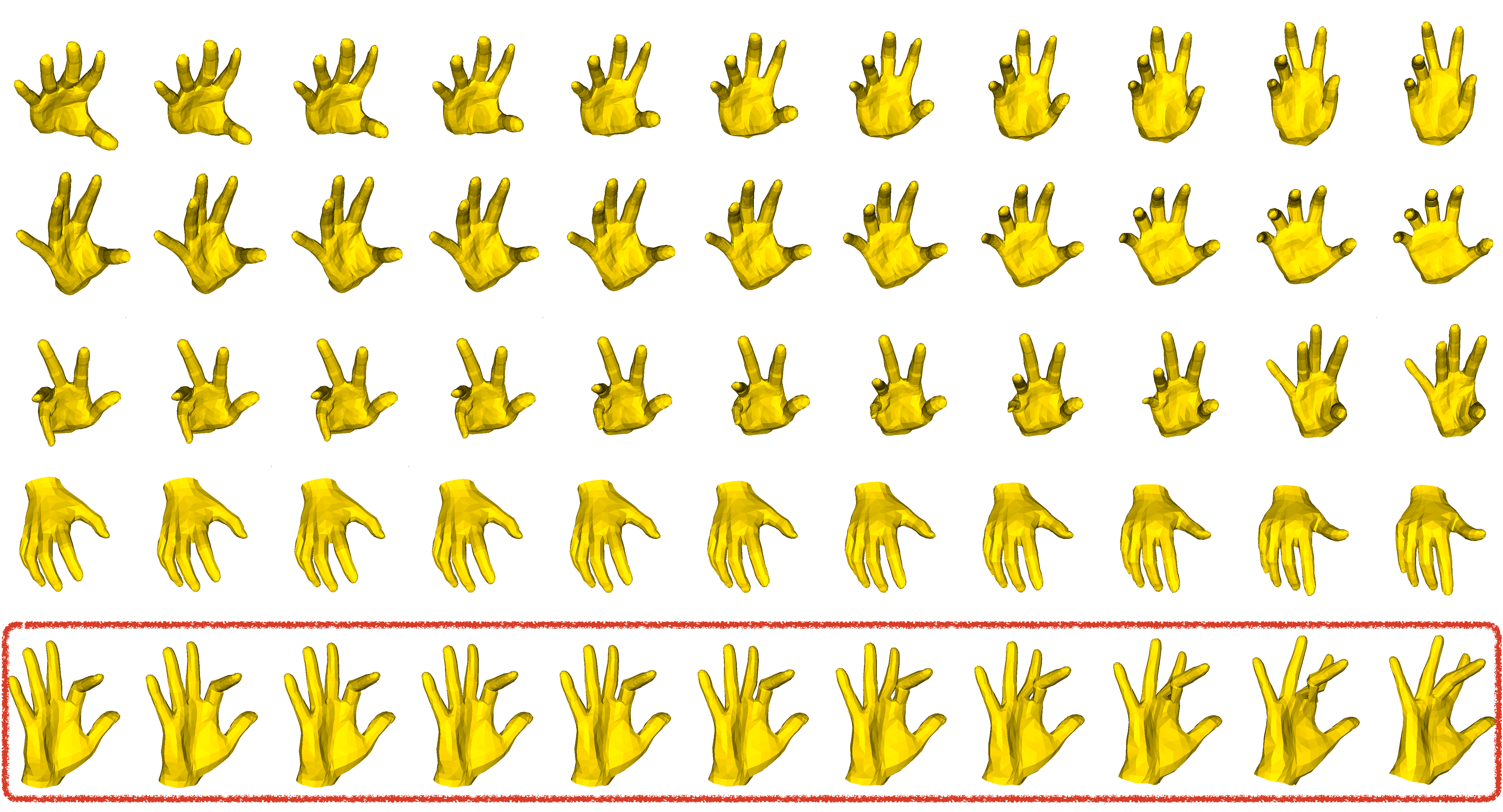}
   \vspace{-0.8cm}
  \caption{Geodesic paths connecting source (left) and target (right) hand shapes modulo deformations. Last row provides and example of a wrong transformation (index and middle fingers interchanged).}
   \label{Fig:ExampleHands}
\end{figure}

-- Following LDDMM algorithms (e.g. \cite{younes2009evolutions} and \cite{miller2006geodesic}), our \textbf{regularizer} is a summation over all kinetic energies of the system at all time-steps ${\cal S}(f)=\frac{1}{2}\int_0^1 {\cal S}_t(f^t) dt$, where, ${\cal S}_t(f)=\| f^t(\Phi^{\Theta}(t,.),\theta^{t}) \|_{\ell_2}^2$. The main difference is that while LDDMM makes use of the $\|.\|_{\cal A}$ of the RKHS induced by Gaussian Kernel $K$ with a fixed deformation scale, our ResNet-LDDMM uses the $\|.\|_{\ell_2}$ on the neural functions $f^l(.,\theta^l)$. Consequently, ResNet-LDDMM generates geodesics by minimizing the amount of energy spent to get close to $q_T$ from $q_S$, when traveling along the Orbit of $q_S$. We summarize the steps of our algorithm ResNet-LDDMM in Algo. \ref{Algo:ResNet-LDDMM}.    

\begin{algorithm}
\caption{ResNet-LDDMM.}
\begin{algorithmic}
\REQUIRE Source shape $q_{S}$, Target shape $q_{T}$, $L$: \# of blocks, $\eta$: learning rate, $\Delta^L = 1/L$: time-step, E:\# of iter., $m$: ResNet's width (\# of filters), $\sigma$ (see Eq. (\ref{Eq:ResNet-LDDMM})) .  
 \STATE $ l \leftarrow 1$, $e \leftarrow 0$, $\Phi^0=I_{\real^3}$
 \STATE Set $ \Theta=\{\theta^l\}_{l \in {1..L}} \leftarrow \Theta_0$ (initial guess)
 \STATE Set $q^{0} \leftarrow \Phi^0.q_S$ (initial state)
 \WHILE{$e<E$ (epoch)} 
  \WHILE{$l<L$ (building block)} 
  \STATE Compute $v^{l} \leftarrow f^{l}(\Phi^{l-1}.q^{l-1},\theta^{l})\Delta^L$
  \STATE Update $\Phi^{l} \leftarrow \Phi^{l-1} + v^{l}$ 
  \ENDWHILE
  \STATE Compute $q^{L}\leftarrow q^0+\sum_{l=1}^{L} v^l$ ($\Phi^{L}\leftarrow \Phi^0+\sum_{l=1}^{L} v^l$)
  \STATE Evaluate the loss function $\mathcal{J}(\Theta^{e})$ (Eq. (\ref{Eq:ResNet-LDDMM})) at epoch $e$
  \STATE Update $\Theta_{e+1} \leftarrow \Theta_{e} - \eta \nabla_{\Theta} \mathcal{J}(\Theta)$ 
\ENDWHILE
\STATE Compute $\Theta^*$ minimizer of Eq. (\ref{Eq:ResNet-LDDMM}) using ADAM.
   \ENSURE $\Phi^*(.,1) \leftarrow \Phi^{L}$ parameterized by $\Theta^*$: final transformation; $\{f^l\}_l$: flow of velocity fields.
 \end{algorithmic} 
\label{Algo:ResNet-LDDMM}
\end{algorithm}

We show in Fig. \ref{Fig:Example} both paths from a source shape (on the left) and a target shape (on the right). We illustrate the flows of velocity fields obtained for $L=10$ intermediate time-steps. We can see also how close are the deformed sources to the target shapes (from the absolute spatial deviations). The examples involving hands, shown in Fig. \ref{Fig:ExampleHands}, are particularly interesting (because of the nature of deformations possible for the hand) and challenging at the same time (as different neighboring fingers can exhibit opposite movements). The last example, depicted in a red box, shows a case where ResNet-LDDMM fails in matching corresponding fingers. In fact, while the transformation is a diffeomorphism (by swapping the index and middle fingers), an incorrect registration result is obtained. Algo. \ref{Algo:GeometricFramework} summarizes the steps for building a geodesic path connecting a source shape to a target shape using the output transformations of Algo. \ref{Algo:ResNet-LDDMM}. Note that here we make use of the EMD distance $\cal D^{EMD}$ as data attachment term, more suitable to compare articulated shapes than the \textit{Chamfer's distance}.

\begin{algorithm}
\caption{Optimal Trajectory on the \textit{Orbit} of $q_S$.}
\begin{algorithmic}[1]
\REQUIRE Source shape $q_{S}$; Target shape $q_{T}$; $\sigma$ (Eq. (\ref{Eq:ResNet-LDDMM})).
\STATE $l \leftarrow 1$ 
\STATE Compute $\Phi(1)$, $\{f^l\}_l$ using Algo. \ref{Algo:ResNet-LDDMM}.
\WHILE{$l\leq L$}
  \STATE Compute $\dot{q}^{l} \leftarrow f^{l}(.,\theta^l)$ ($\dot{q}^{l} \in T_{q^l}([q_S])$: the instantaneous shape's speed, $[q_S]$: Shape Space)
  \STATE Update the shape $q^{l+1} \leftarrow q^{l} + \dot{q}^{l}$ (on $\cal S$)
  \ENDWHILE
\ENSURE Discrete steps $q^{l}$, ${l\in\{1,\dots,L\}}$ of an optimal trajectory connecting $q^{0} \leftarrow q_S$ to $q^{L}\sim q_T$.
\end{algorithmic} 
\label{Algo:GeometricFramework}
\end{algorithm}

\begin{figure*}[ht!]
  \centering
\includegraphics[width=\linewidth]{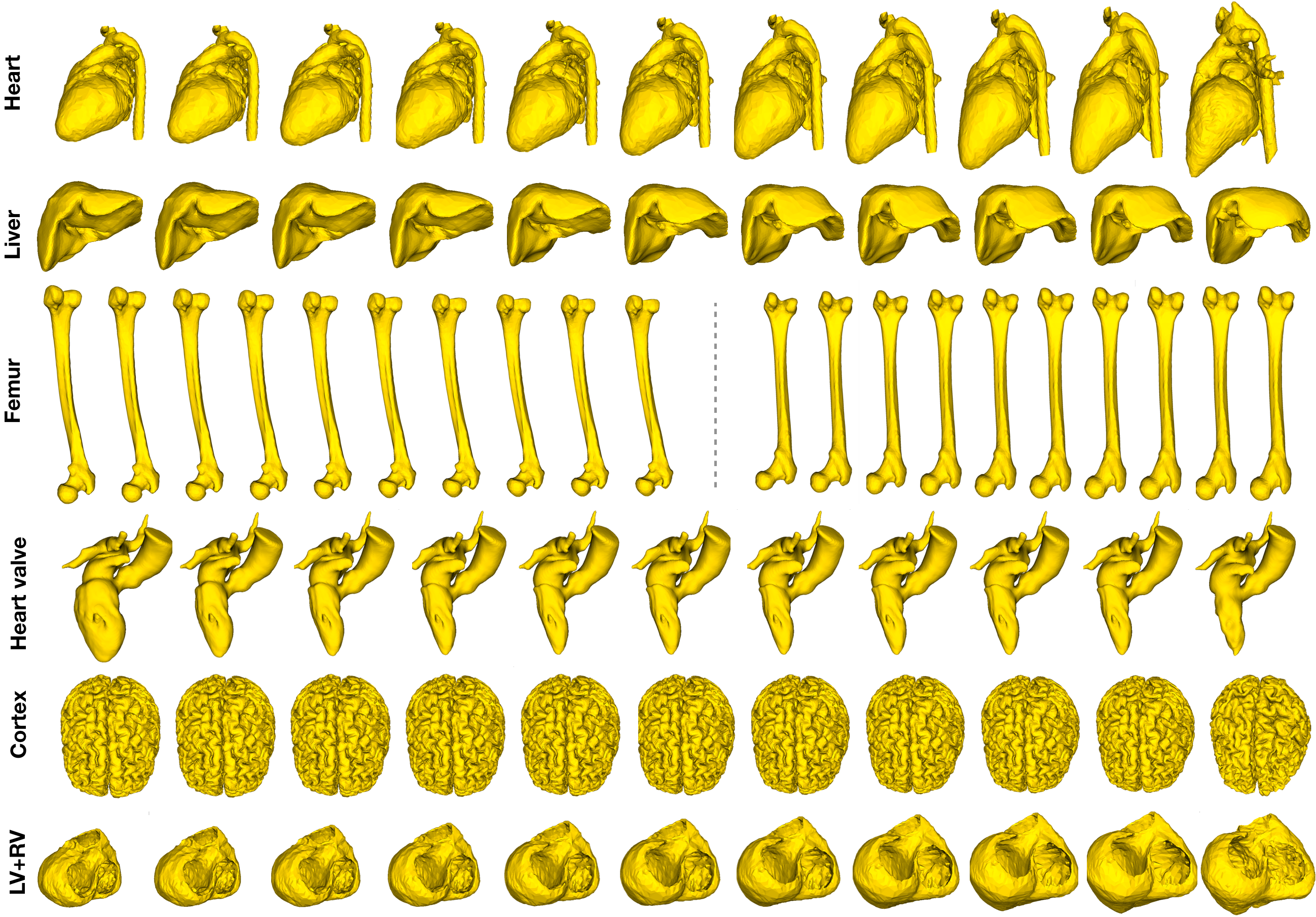}
    \vspace{-0.5cm}
  \caption{Geodesic paths connecting source (first column) and target anatomical shapes (last column) of different subjects. From top to bottom: whole heart, liver, femur 1-to-2 (left) followed by femur 2-to-1 (right), heart valve, cortex and LV+RV (left and right ventricles of pathological hearts).}
   \label{Fig:CAExamples}
\end{figure*}

To demonstrate the performance of ResNet-LDDMM in registering anatomical parts of the human body, we provide in Fig. \ref{Fig:CAExamples} several examples of optimal deformations. Shapes are 3D triangulated surfaces obtained from manual segmentation of MRI images of different subjects. Note that here the $\cal D^{CD}$ distance is used to compute the data attachment term in Eq.(\ref{Eq:ResNet-LDDMM}). In each row, we show the source shape on the left, the target shape on the right and discrete steps from the geodesic path connecting them, except for the femur example, where both paths from the source to the target and inversely, are reported. 

\section{Dynamic and Geometric Analysis of ResNet-LDDMM, and How it Builds Diffeomorphisms}
\label{sec:Diffeos}

\subsection{Regularity of the Deformations}
Each building block $f(\cdot,\theta^l)$ of ResNet-LDDMM can be seen as a vector field on $\real^3$, that is, a mapping $\real^3\rightarrow\real^3$. This yields the transformations $\Phi^l:\real^3\rightarrow\real^3$
\begin{equation} \label{Eq:discreteDiffeo}
\begin{split}
&\forall x\in\real^3,\ l\in \{1,\dots,L\},\\ &\Phi^0(x)=x, \\&\Phi^l(x)=\Phi^{l-1}(x)+ f(\Phi^{l-1}(x),\theta^l).\Delta^L.
\end{split}
\end{equation}
Then, applying $\Phi^l$ separately to each point of the source shape $q_0$, we get the deformed shapes $q^l=\Phi^l.q_0$. Moreover, for a given $x\in q_0$, each $f^l$ is actually computed explicitly (using Eq. (\ref{Eq:vectorFields})) by,
\begin{equation} \label{Eq:vectorFields}
f(x,\theta^l) = w_3^l(w_2^l((w_1^l x+b_1^l)^++b_2^l)),
\end{equation}
with $w_i^l$ matrices whose entries are made up of the parameters $\theta^l$, along with the real numbers $b_j^l$, and $r^+=\max(r,0)=ReLU(r)$, for every real number $r$, is the ReLU function. Each $f(\cdot,\theta^l)$ is Lipshitz, with Lipshitz constant at most equal to the product of the matrix norms of $w_1^l,w_2^l$ and $w_3^l$. The fact that the corresponding transformation is a diffeomorphism can mostly be deduced from  \cite{rousseau2020residual}, but because of minor differences, we give a quick summary of a proof in the following (with details in the Appendix). The system (\ref{Eq:discreteDiffeo}) is a forward Euler algorithm for the flow of the ODE $\partial_t\phi(t,x)=f(x,\theta(t))$, for $x$ in $\real^3$, $t$ in $[0,1]$, and $\theta(t)=\theta^l$ on the interval $[(l-1)\Delta L,l\Delta L)$. Hence, $\phi$ is the flow of the time-dependent vector field $f$, whose Lipshitz constant in space is always less than the maximum $\mathcal{M}$ over $l$ of the product of the matrix norms of $w_1^l,w_2^l$ and $w_3^l$. Therefore, $\phi(t):\real^3\rightarrow\real^3$ exists for all time $t$ in [0,1], and is a bilipshitz map for all $t$. Grownall's lemma also ensures that its Lipshitz constant and that of its inverse is less than $e^{t\mathcal{M}}$. As a final note, should smoothness be required, the ReLu can be replaced by a smooth approximation $g$ with bounded derivative. In this case, the velocity fields will be as smooth as $g$ and the transformation $\Phi$ will then truly be a diffeomorphism. However, as shown in our various experiments, this does not appear necessary, and doing so intrinsically causes the loss of one of the strong points of our approach compared to classical LDDMM: the absence of a scale (see Sec. \ref{sec:LDDMMcomparison}).

\subsection{Geometric Description of the Velocity Fields and Interpretation}
\label{sec:GeometricDescription}

For fixed $l$, the vector fields $f^l(\cdot,\theta^l)$ given by the $l$-th building block have a nice geometric interpretation, which helps understand the way ResNet-LDDMM functions work and the areas in which it improves upon some of the classical LDDMM's weaknesses. For readability, we will omit the index $l$ for now. Recall that $m$ denotes the width of the network. For a fixed parameter $\theta$ of a building block $f(\cdot,\theta)$ of our network  (i.e., a velocity field $x\mapsto f(x,\theta) $), the first layer is the affine transformation $L_1:x\in\real^3\mapsto w_1x+b_1\in \real^m$, with $w_1$ being an $m\times 3$ matrix and $b_1=(b_{1,1},\dots,b_{1,m})$ in $\real^m$. Let us now note $n_1,\dots, n_m\in \real^3$ the vectors whose transposes are the lines of $w_1$; that is, $w_1=(n_1,\dots,n_m)^T$. Then $L_1(x)^T=(n_1^Tx+b_{1,1},\dots,n_m^Tx+b_{1,m})$. 

The second layer $L_2:\real^m\rightarrow\real^m$ is a ReLU function, so that
\begin{equation}\begin{split}
    L_2\circ L_1(x)^T=&((L_2\circ L_{1})_1(x),\dots,(L_{2}\circ L_1)_m(x))
    \\=&((n_1^Tx+b_{1,1})^+,\dots,(n_m^Tx+b_{1,m})^+)
    \end{split}
\end{equation}

Geometrically, a vector $n\in \real^3$ and a number $b\in \real$ define an affine plane $P$ of $\real^3$ through the equation $n^Tx+b=0,\ x\in\real^3$. This separates $\real^3$ into two half spaces: $E^+=\{x\in\real^3, \ n^Tx+b\geq 0\}$ on the side of $P$ towards which $n$ is pointing, and $E^-=\{x\in\real^3, \ n^Tx+b<0\}$, on the other side of $P$. Then, for any $x$ in $\real^3$, we have that the function

\begin{equation}\begin{split}
    D_{n,b}(x)=&(n^Tx+b)^+\\
    =&\left\lbrace\begin{split}n^Tx+b\ \text{if}\ x\in E^+,\\
    0 \text{ if}\ x\in E^-.\end{split}\right.
    \end{split}
\end{equation}
Then we notice that for $k=1,\dots, m$, the k-th component of $L_2\circ L_1$ is just $(L_2\circ L_1)_k=D_{n_k,b_k}$. 

Hence, the choice of parameters for the first two layers is equivalent to choosing $m$ oriented affine planes $P_1,\dots,P_m$ with corresponding Cartesian equations $n_k^Tx+b_k=0$, $k=1,\dots,m$, and half-spaces $E_k^{\pm}$, obtaining the corresponding components $D_k=D_{n_k,b_k}$. 
The third and fourth layers of the building block simply combine into an affine transformation $\real^m\rightarrow \real^3$. Therefore, each of the three components of the entire building block $f(.,\theta):\real^3\rightarrow\real^3$ is just an affine combination of the functions $D_k$. In other words, for some vectors $a_1,\dots,a_m$ and $c$ in $\real^3$, we have
\begin{equation}\label{eq:GeometricVelocityFields}
    f(x,\theta)=a_1D_1(x)+\dots+a_mD_m(x)+c.
\end{equation}

Now, take a finite sequence of symbols $\epsilon = (\epsilon_1,\dots,\epsilon_m)$ with each $\epsilon_k$ being either $+$ or $-$, and consider the subset $\displaystyle T_\epsilon=\cap_{k=1,\dots,m} E_k^{\epsilon_k}.$ The domain $T_\epsilon$ is a polytope, that is, a possibly unbounded polyhedron. Then, for every $k$ such that $\epsilon_k=-$, $D_k(x)=0$, and for every other $k$, $D_k(x)=n_k^Tx+b_k$. Plugging this formula into Eq.(\ref{eq:GeometricVelocityFields}), we get
\begin{equation}\label{eq:GeometricVelocityFieldsT}
\begin{split}
    f(x,\theta)=&\sum_{k,\epsilon_k=+}(\underset{\text{3 by 3 matrices}}{\underbrace{a_{kn_k^T}}x}+b_k)+c
    \\
    =&\left(\sum_{k,\epsilon_k=+}a_{kn_k^T}\right)x+\sum_{k,\epsilon_k=+}b_k+c
    \\
    =&A_{\epsilon}x+c_\epsilon.
\end{split}
\end{equation}
In other words, on each $T_\epsilon$, $f(\cdot,\theta)$ is just an affine vector field. In conclusion, each building block of our network, that is, each velocity field we use to construct our final transformation, is just a globally continuous (even globally lipshitz), piece-wise affine mapping, defined on a partition by $2^m$ disjoint polytopes of $\R^3$. 

\begin{figure} [ht!]
  \centering
  \includegraphics[width=\linewidth]{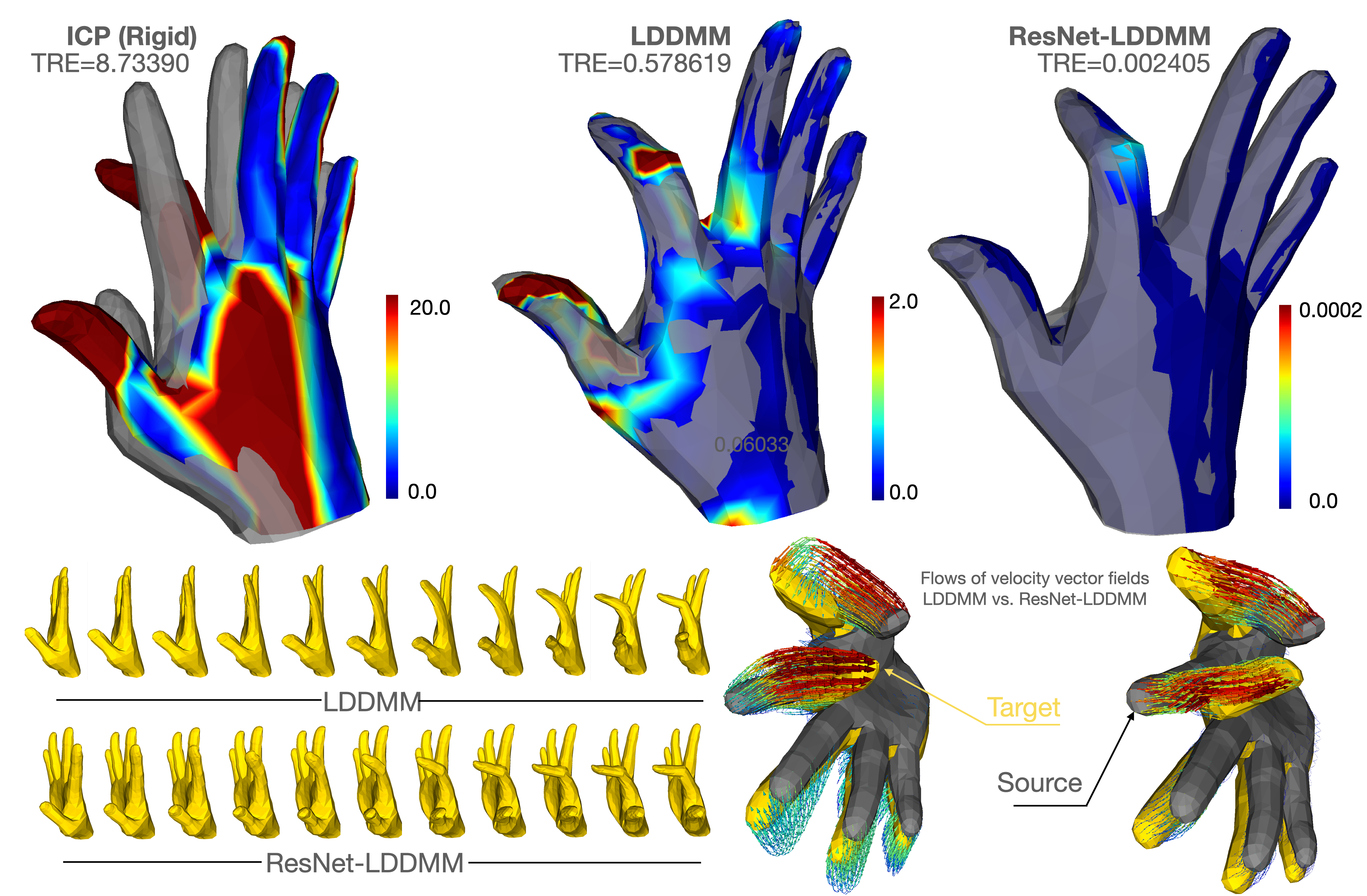}
  \caption{LDDMM vs. ResNet-LDDMM -- registration results of hand shapes with fingers moving in opposite directions.}
   \label{Fig:Directions}
\end{figure}

The optimization of the final network, ResNet-LDDMM, can therefore be thought of as searching for:
\begin{enumerate}
    \item The optimal partition of $\real^3$ into polytopes by $m$ planes at each step
    \item The correct affine transformation on each of these polytopes.
\end{enumerate}
While there are obviously some additional constraints (for example, the transformations need to be globally continuous), this description gives a good representation of what ResNet-LDDMM does. As a result, different parts of the shape, even if they are close in $\real^3$ (such as two fingers on a hand) can be made to belong to different polytopes, and therefore be moved in different directions without costing too much energy (see Fig. \ref{Fig:Directions}). 

\subsection{Regularization Term}
\label{sec:GeometricRegularization}

Let us first give a reason not to use (or at least not to only use) a more classical regularization term on the parameters $\theta^l$. Keeping in mind the dynamical interpretation of ResNet-LDDMM, such a term would be given by $\frac{1}{L}\sum_{l=1}^L\Vert\theta^l\Vert^2_{2}$, possibly multiplied by a weight, and corresponds to $\int_0^1\Vert\theta(t)\Vert^2_{2}$dt in the continuous model. However, since each residual layer $f(\cdot,\theta^l)$ is, piecewise, a polynomial of degree three with in the parameters of $\theta^l$, a regularization on $\theta^l$ does not give a good control on its Lipschitz constant. In the network itself, that is not too much of a problem, since boundedness is still ensured. However, in the continuous model, it would be possible to have parameters $\theta(t)$ that have bounded $L^2$ norm, but induce velocity $f(\cdot,\theta(t))$ whose Lipshitz constant $k(t)$ is not integrable. This would prevent it from integrating  into a well-defined diffeomorphism. Indeed, just take each coordinate of $\theta(t)$ to be $\frac{1}{t^{1/3}}$. Then $\Vert\theta(t)\Vert^2_{2}\simeq \frac{1}{t^{2/3}}$ is integrable, but the Lipshitz constant $k(t)\simeq\theta(t)^3\simeq\frac{1}{t}$ is not. This is a known problem in optimal control and shape analysis, and almost always prevents the existence of a global minimizer. See \cite{mumford2006riemannian} for example.

Instead of a regularization term on the parameters $\theta^l$, we use the sum of the kinetic energies of the paths taken by each point of the shape along the deformation \eqref{Eq:ResNet-LDDMM}, which is equal to
$$
\frac{1}{2}\sum_{i=1}^n\int_0^1\Vert f(\Phi(x_i,t),\theta^t)\Vert_{\ell^2}^2dt.
$$
If we had enough degrees on freedom on $\theta$, the optimal deformation $\Phi^*$ with parameters $\Theta^*$ would then be such that every  curve $t\in[0,1]\mapsto \Phi(x_i,t)$ is a straight segment of constant speed, since those minimize the kinetic energy, making the whole algorithm rather pointless. In LDDMM, this happens when the scale becomes much too small. For ResNet-LDDMM, it would correspond to a network width that is too large, allowing each point to belong to a single polytope  with complete control on the affine transformation inside this polytope (as described in the previous section, Section \ref{sec:GeometricDescription}). However, for smaller width, at each residual layer $l$, each polytope $T$ will contain several points of the shape. Then, as long as four those points are not co-planar, the affine mapping $A$ induced by $f(\cdot,\theta^l)$ on $T$ is completely described by its value at those points. As a result, the lipshitz norm of $A$ is indeed controlled by the Euclidean norm of the values of $A$ at those points.

\section{Implementation and Experiments}
\label{sec:Experiments}

In this section, we discuss both qualitative and quantitative registration results of our ResNet-LDDMM on different datasets. We compare some of our results to those obtained using a Varifold-type LDDMM implementation \cite{charon2013varifold}. Next, we report evaluations and comparisons on the most recent SHREC datasets (SHREC'2019 in Sec.\ref{Sec:SHREC2019} and SHREC'2020 in Sec.\ref{Sec:SHREC2020}) and discuss strengths and limitations (Sec.\ref{Sec:limitations}). In the supplementary materials (SM), we discuss the influence of the regularization term, the impact of the network width, i.e. the parameter $m$ and the key role that the \textit{ReLU} activation function plays inside the building blocks $f^l$, as ablative studies. We compare the computational complexity of ResNet-LDDMM compared to LDDMM and existing approaches.  We also study the robustness of ResNet-LDDMM to noise and missing data.         

\subsection{Qualitative Comparison to LDDMM}
\label{sec:LDDMMcomparison}

Both LDDMM and ResNet-LDDMM produce a topology-preserving transformation (i.e. a diffeomorphism) of the whole ambient space $\real^3$. However, there are several differences between the two make the ResNet approach both easier to run and more adapted to certain problems, particularly those involving very distinct motions in different areas of the considered shape. This is because LDDMM methods require choosing a certain ``scale". For a discrete surface, when using a Gaussian kernel $K(x,y)=\exp(\frac{\Vert x-y\Vert_2}{2\mu^2})$, the scale is $\mu$, and every vector field generated will be a sum of Gaussian vector fields with fixed variance $\mu$, each centered at a point on the shape (see \cite{arguillere2016multipleshapes} for example). As a result, moving points within distances less than $\mu$ in different directions requires a lot of energy, so such movements generally do not appear when minimizing the functional Eq. (\ref{Eq:LDDMM}). However, the scale cannot be too small, or the motion will no longer need to preserve the topology of the triangulated surface. As a result, two main difficulties appear in LDDMM that the ResNet version improves upon. 

\begin{figure}[ht!]
  \centering
   \includegraphics[width=1\linewidth]{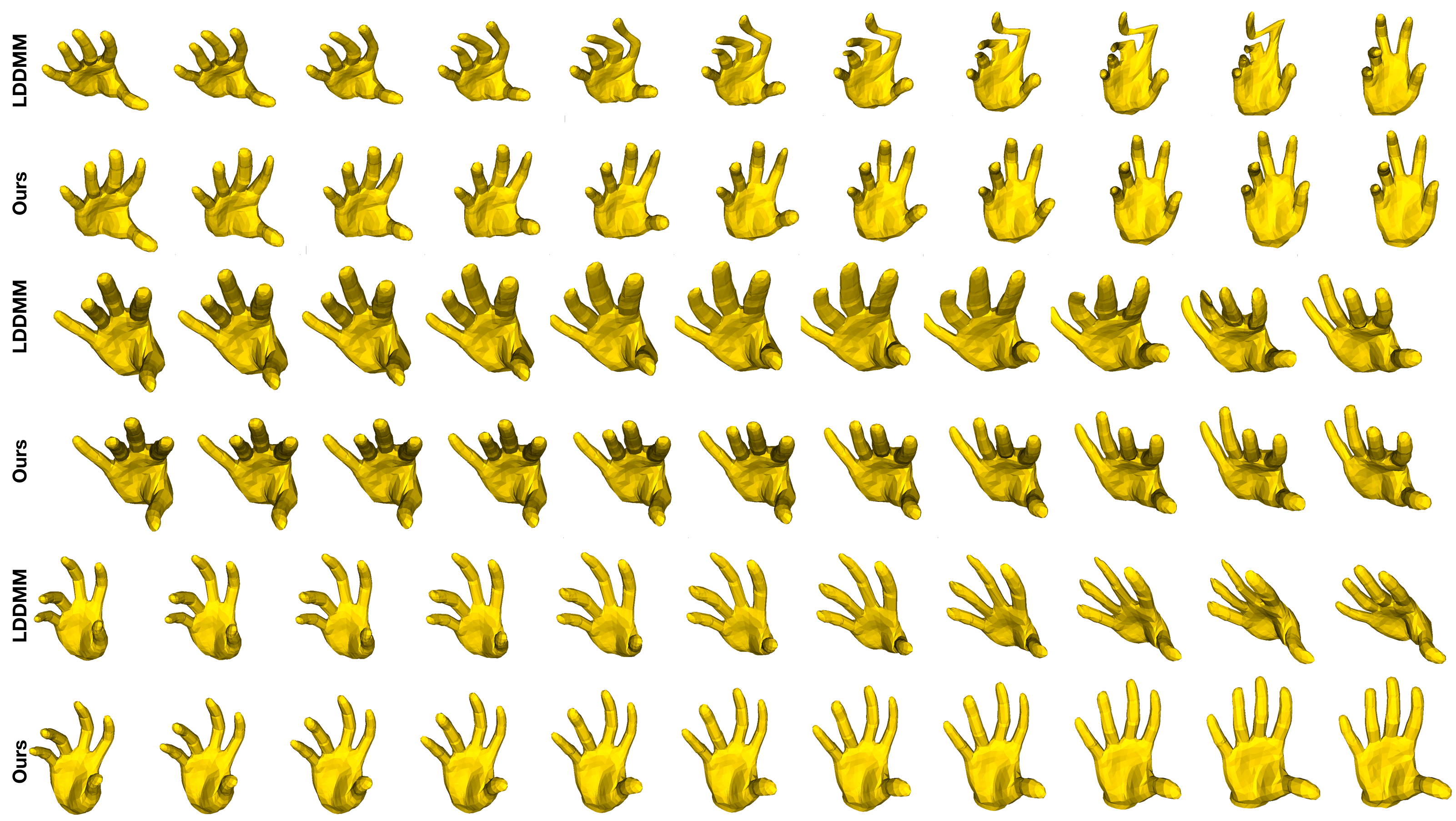}
    \includegraphics[width=\linewidth]{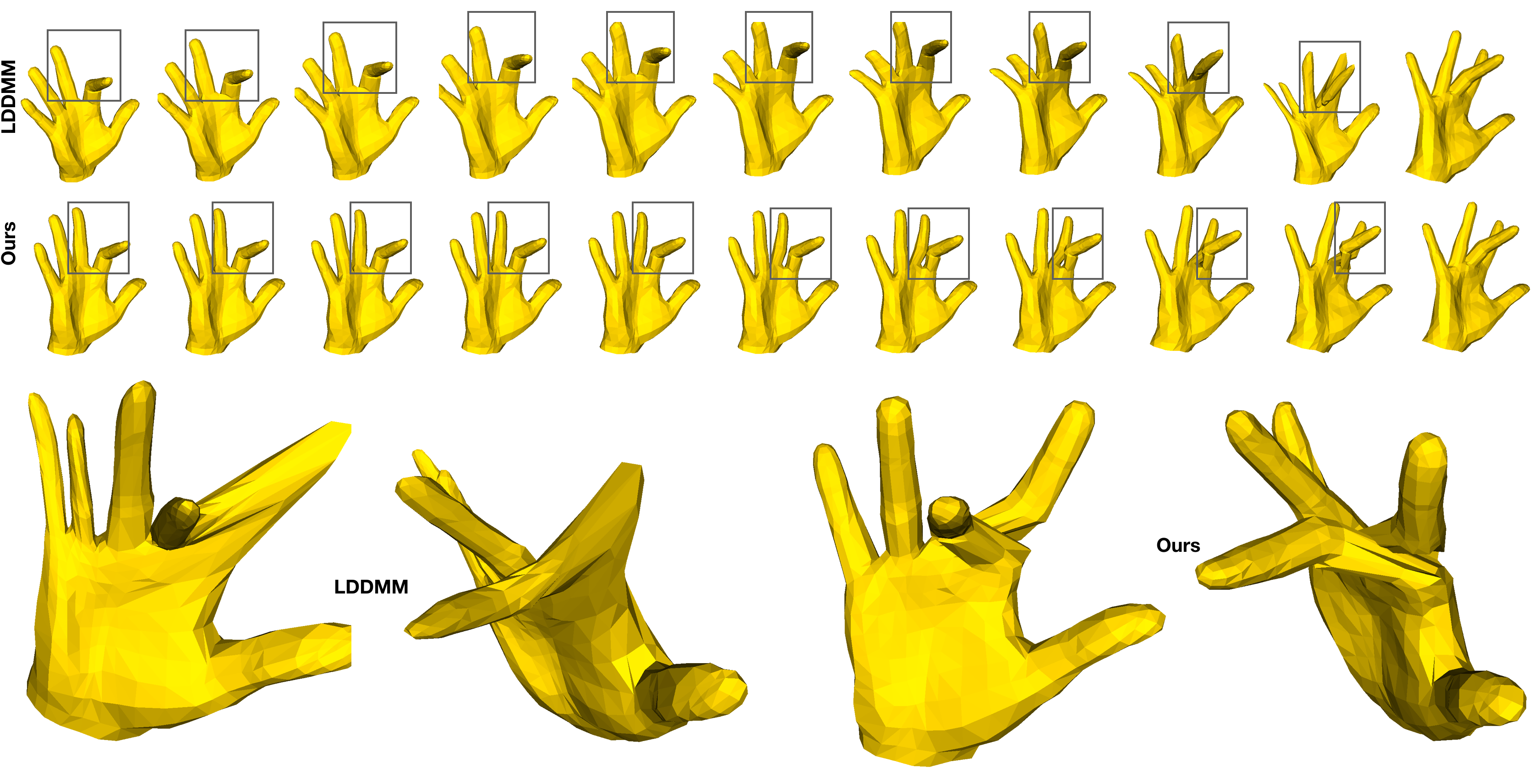}
  \caption{Four examples to compare ResNet-LDDMM (ours) to LDDMM. }
   \label{Fig:ComparisonToLDDMM}
\end{figure}

\textbf{--} First, in LDDMM, one must find a ``good" scale $\mu$, which cannot be too small or too large, in order to allow a wide enough range of motions while still preserving the smoothness of the deformation. That is much less of a problem with ResNet-LDDMM: the size of the polytopes on which the velocity fields are computed not fixed, and so the algorithm automatically computes polytopes and polyhedrons of appropriate sizes. The role of scale falls instead on the width $m$ of the network. However, we will see that the results are not particularly sensitive to a change of width less than an order of magnitude, and simply choosing a width of a few hundred yielded good results on every case we tested. Further there is no automated way to find the best scale, so finding the correct one for LDDMM simply requires testing several values (with a bit of intuition to avoid bad ones). Hence, finding the best scale can sometimes take several days, and so we often settle one that is simply ``good enough". Whether the results of Tab. 2 of the Supplementary Materials are due to not finding the correct scale, or to our ResNet beating the best scale is up to debate, but regardless the ResNet results were obtained on the first try, which for practical applications is a clear improvement. 

\textbf{--} Second, there may not even be a good scale in the LDDMM framework to compare certain shapes. This is generally caused by points belonging to very different parts of the shape that also happen to be very close, such as two fingers on a hand. Such matching problems are generally much harder for the usual LDDMM, because one needs to choose big enough scales of deformation in the reproducing kernel to ensure regularity of the shape. This can either prevent necessary deformations from occurring, or force the transformation to first separate and unnaturally expand the fingers during the first half of the transformation. This is also a well-known problem when tackling multiple shapes (see \cite{arguillere2016multipleshapes} for example). This problem is particularly obvious in Fig. \ref{Fig:ComparisonToLDDMM} when matching various hand positions, especially (but not only) in the intermediate deformation steps. On the other hand, thanks in part to only requiring bi-lipshitz regularity, ResNet-LDDMM can move two parts of the shape (such as two fingers) in opposite directions while keeping the regularization term small by encasing each part within distinct polytopes. Fig. \ref{Fig:Directions} illustrates how ResNet-LDDMM performs better than LDDMM for the case of a hand.

\textbf{--} As a final difference between the methods, LDDMM is quadratic with respect to the resolution (i.e., the number of points), and is notoriously time-consuming to run. While matching two shapes usually takes a  acceptable amount of time, matching hundreds (or even thousands) for adequate statistical analysis can quickly become impossible. Significant advances have been made using GPUs and the free library KeOps (www.kernel-operations.io) on this front \cite{charlier2020kernel}, but it is still time-consuming to perform such an analysis, especially when one needs several retries to find the correct scales for the reproducing kernel. On the other hand, ResNet-LDDMM is much faster (refer to Supplementary Materials), especially at high resolution; the network itself actually has linear complexity with respect to the resolution. Only the data attachment terms have quadratic complexity. Combined with the lack of needing compute a correct scale, this new method seems like a clear winner in this area (at least for more than a few hundred points).

\subsection{Experiments on SHREC'2019}
\label{Sec:SHREC2019}

The SHREC'2019 dataset (Task.8) consists of articulated wooden mannequins and wooden hands. The organizers have created clothes for the models from two materials; one that can bend but is resistant to stretching, and another that can bend and stretch. To induce greater non-isometry, they used plasticine underneath the clothing of the model \cite{dyke2019shrec}. Because the dataset consists of real-world scans, it contains geometric inconsistencies and topological change caused by self-contact. The real-scans also contain natural noise, varying triangulation of shapes and occluded geometry. Four test sets have been released as follows: Test-set0 contains 14 pairs of articulating wooden hand objects; Test-set1 contains 26 pairs of models, comprising clothed humans and hands (limited to near-isometric deformations); Test-set2 contains 19 pairs of models; the pairings are between a thin clothed mannequin and a larger mannequin, ensuring significant non-isometry; Test-set3 contains 17 carefully selected pairs that contain challenging geometric and topological changes. 

\begin{figure}[ht!]
  \centering
   \includegraphics[width=1\linewidth]{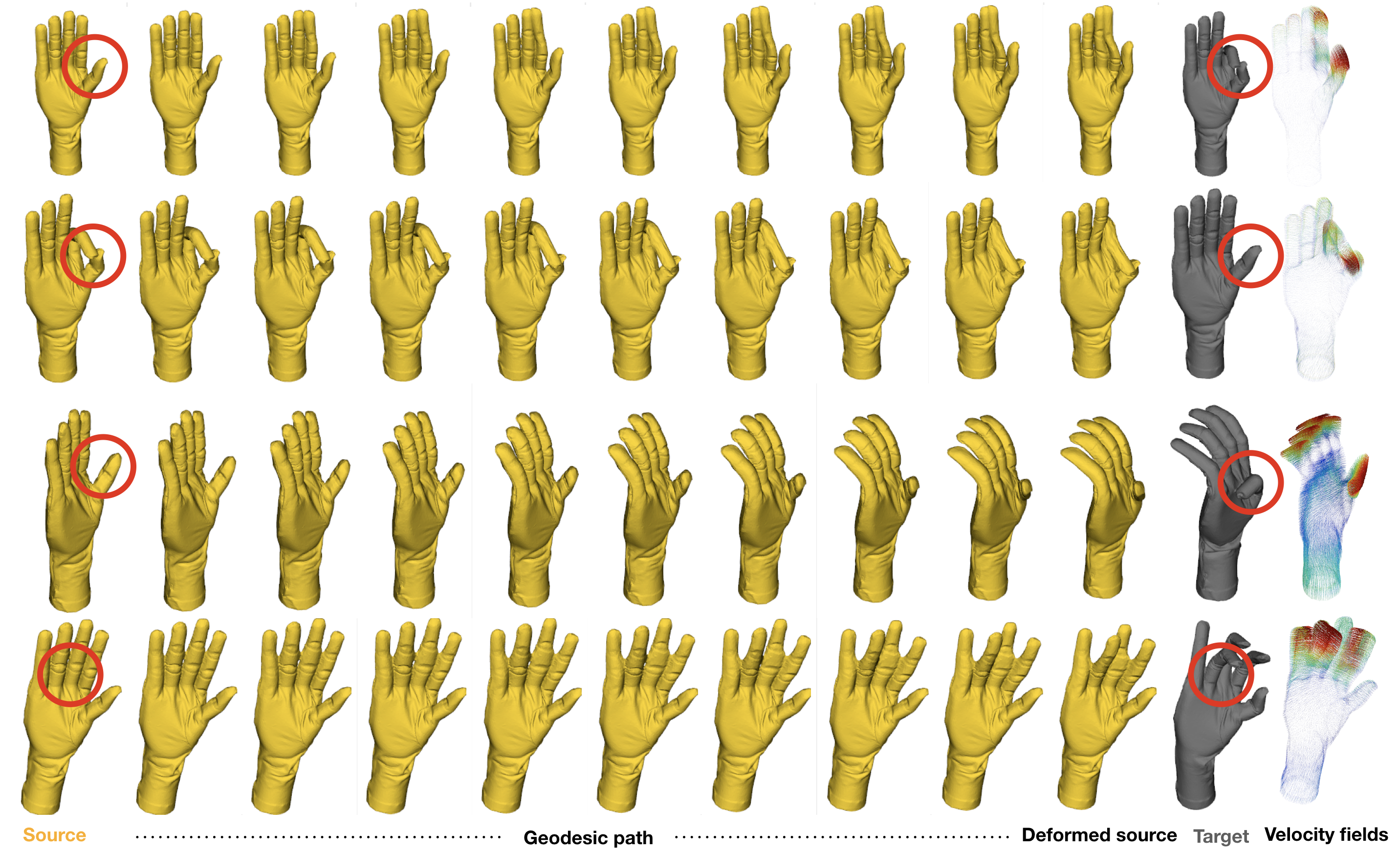}
  \caption{Deformable registration achieved by ResNet-LDDMM (geodesic paths and computed velocity fields): Difficult examples where source and target shapes have different topology (highlighted in red circles).}
   \label{Fig:TopologyExamples}
\end{figure}

\begin{figure*}[ht!]
  \centering
   \includegraphics[width=.19\linewidth]{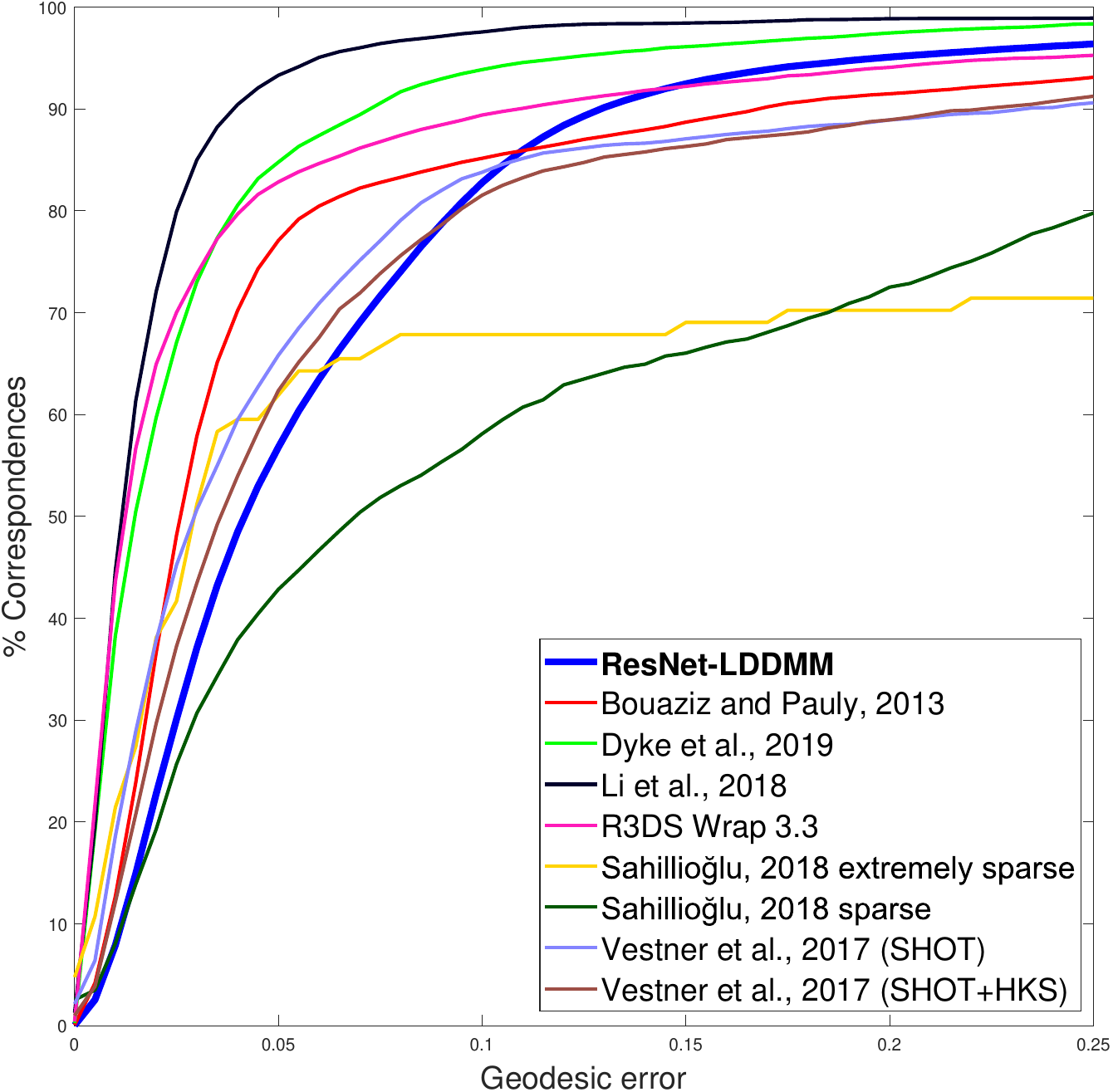}
   \includegraphics[width=.19\linewidth]{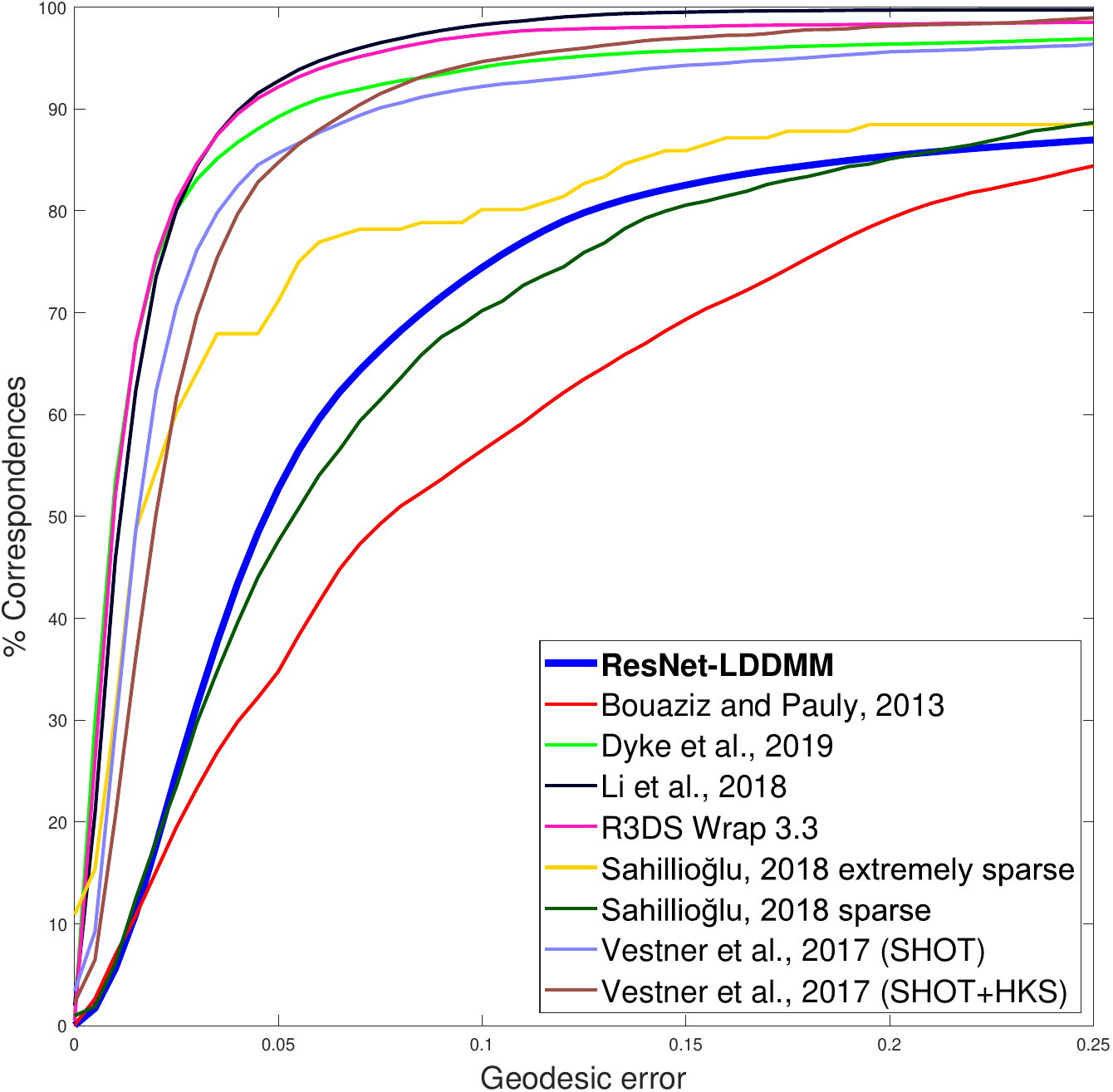}
    \includegraphics[width=.19\linewidth]{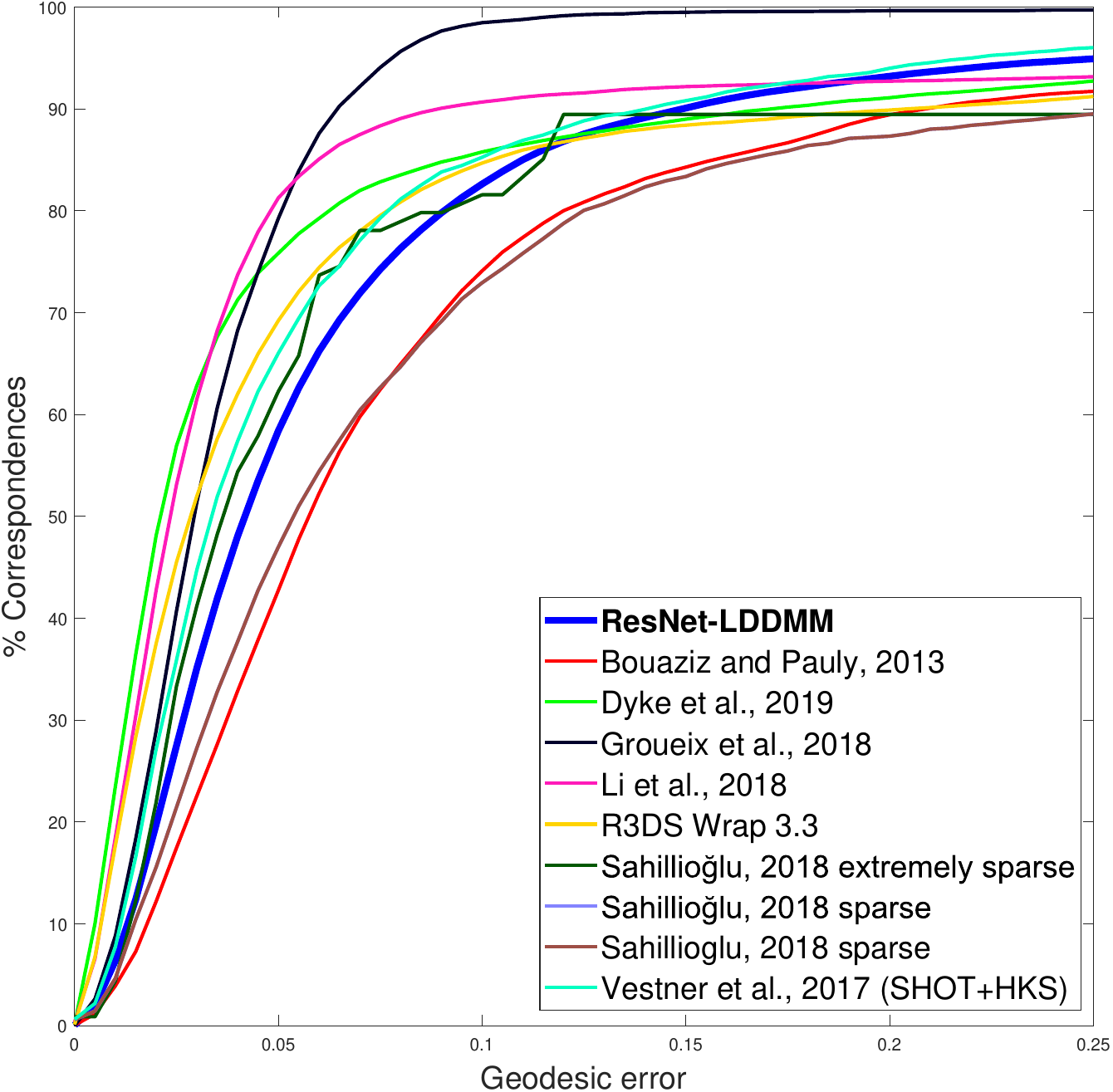}
   \includegraphics[width=.19\linewidth]{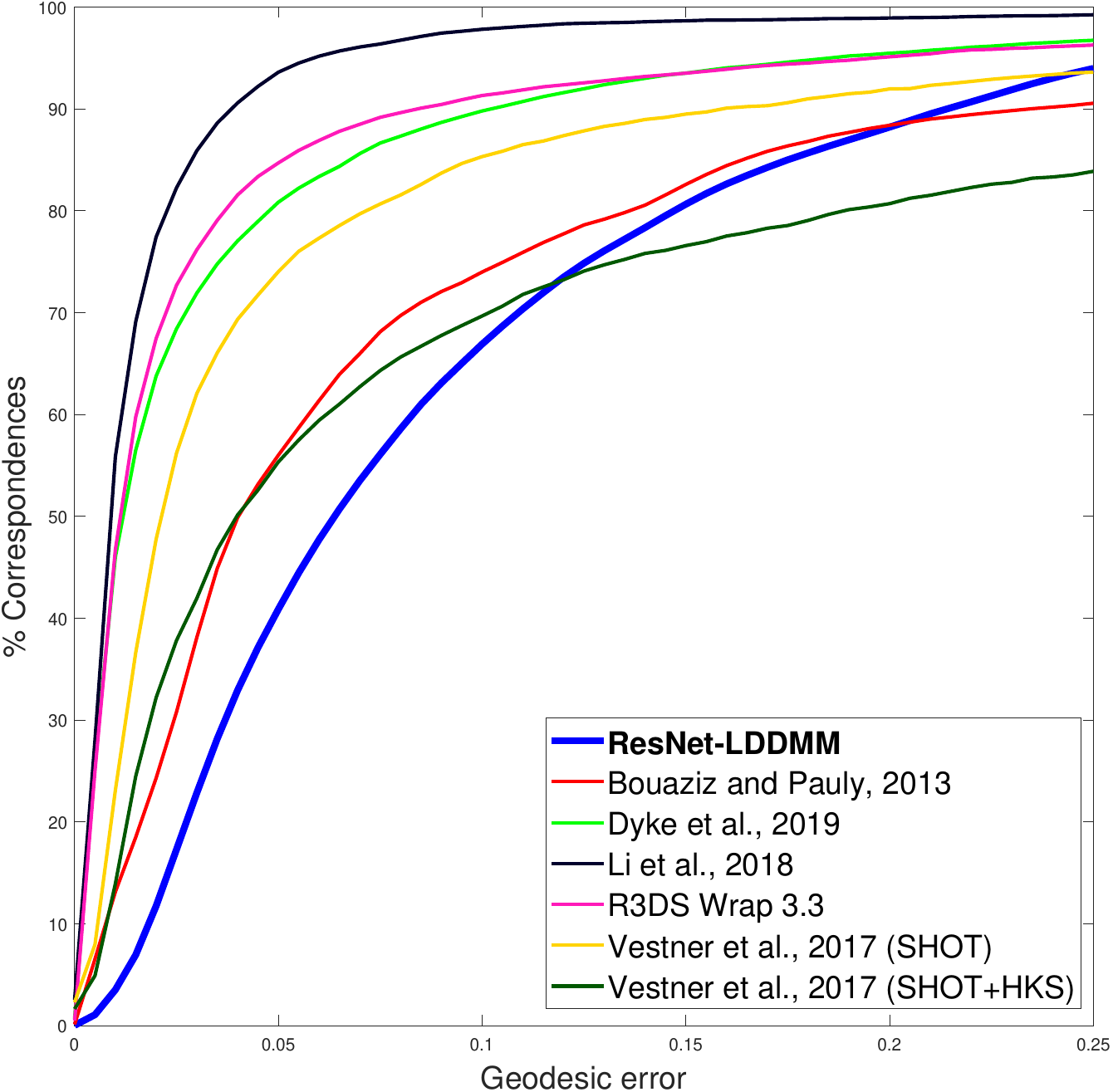}
      \includegraphics[width=.19\linewidth]{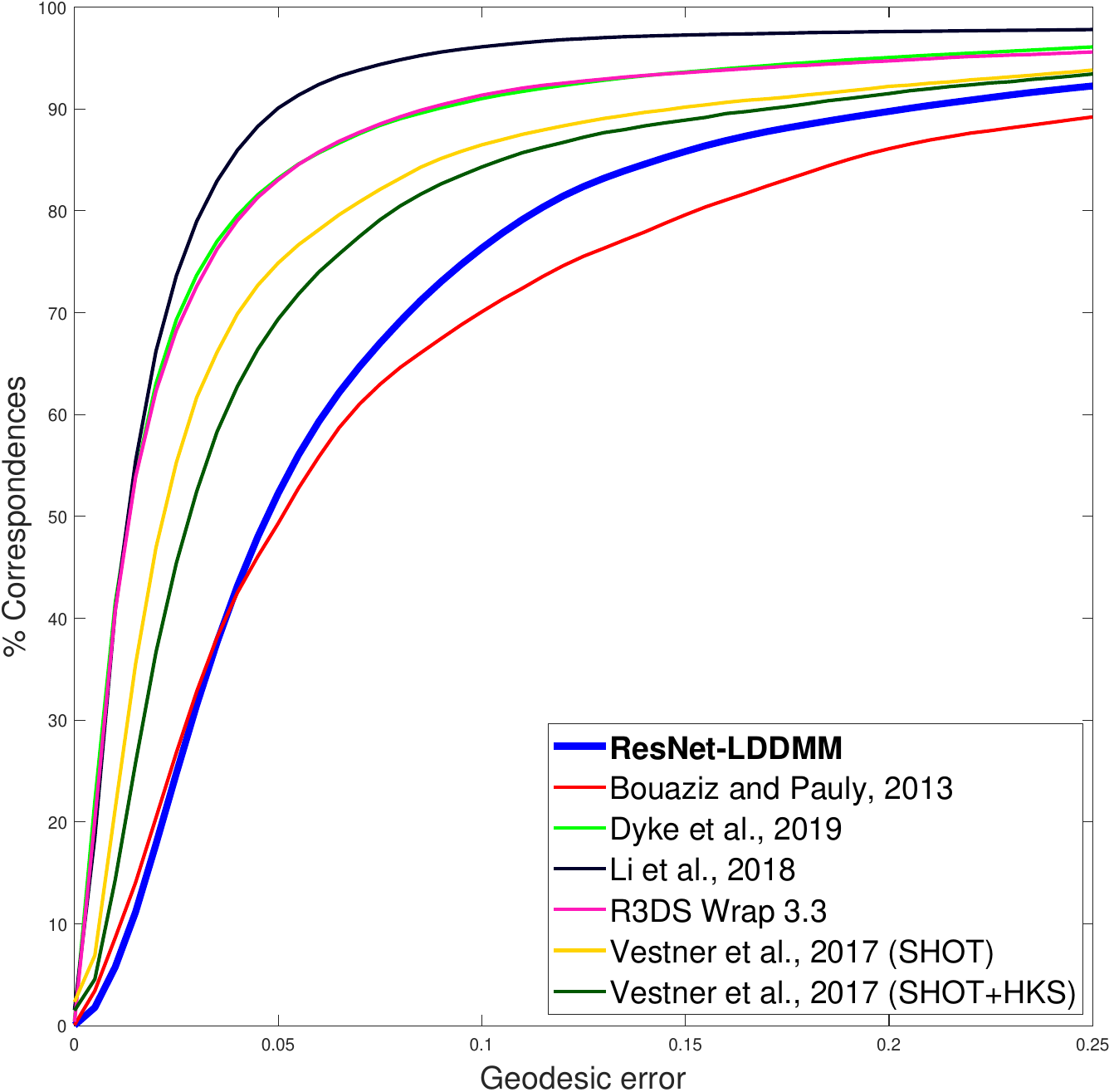}
        \vspace{-0.3cm}
\caption{Evaluation on SHREC'2019 and comparison with existing methods (from \cite{dyke2019shrec}). From left to right : test-set0 (articulated deformations), test-set1 (near-isometric deformations), test-set2 (non-isometric deformations), test-set3 (topological and geometric changes), all test-sets.}
   \label{Fig:SHREC2019}
\end{figure*}

Results and comparison with existing methods are reported in Fig.\ref{Fig:SHREC2019}. We notice that only Dyke \etal \cite{dyke2019non} (and ours) have reported dense registration results. Also, while the majority of approaches start from an initial correspondence (based on SHOT descriptors) along with the diffusion pruning framework, our approach doesn't require such initialization. Finally, unlike the 3D-CODED approach \cite{groueix20183d}, tested only on test-set2 (human-like articulated mannequin), which requires a huge amount of data to train the network, ResNet-LDDMM is completely unsupervised. It is clear from these results (the fourth column, results on test-set3, reflects clearly this issue) the limitation of our ResNet-LDDMM to handle topological changes between the source and target shape. That is expected as, following LDDMM, ResNet-LDDMM was designed to compute a topology-preserving transformation. Examples reported in Fig.\ref{Fig:TopologyExamples} illustrate this limitation of our ResNet-LDDMM. 

\begin{figure*}[ht!]
  \centering
    \includegraphics[width=0.19\linewidth]{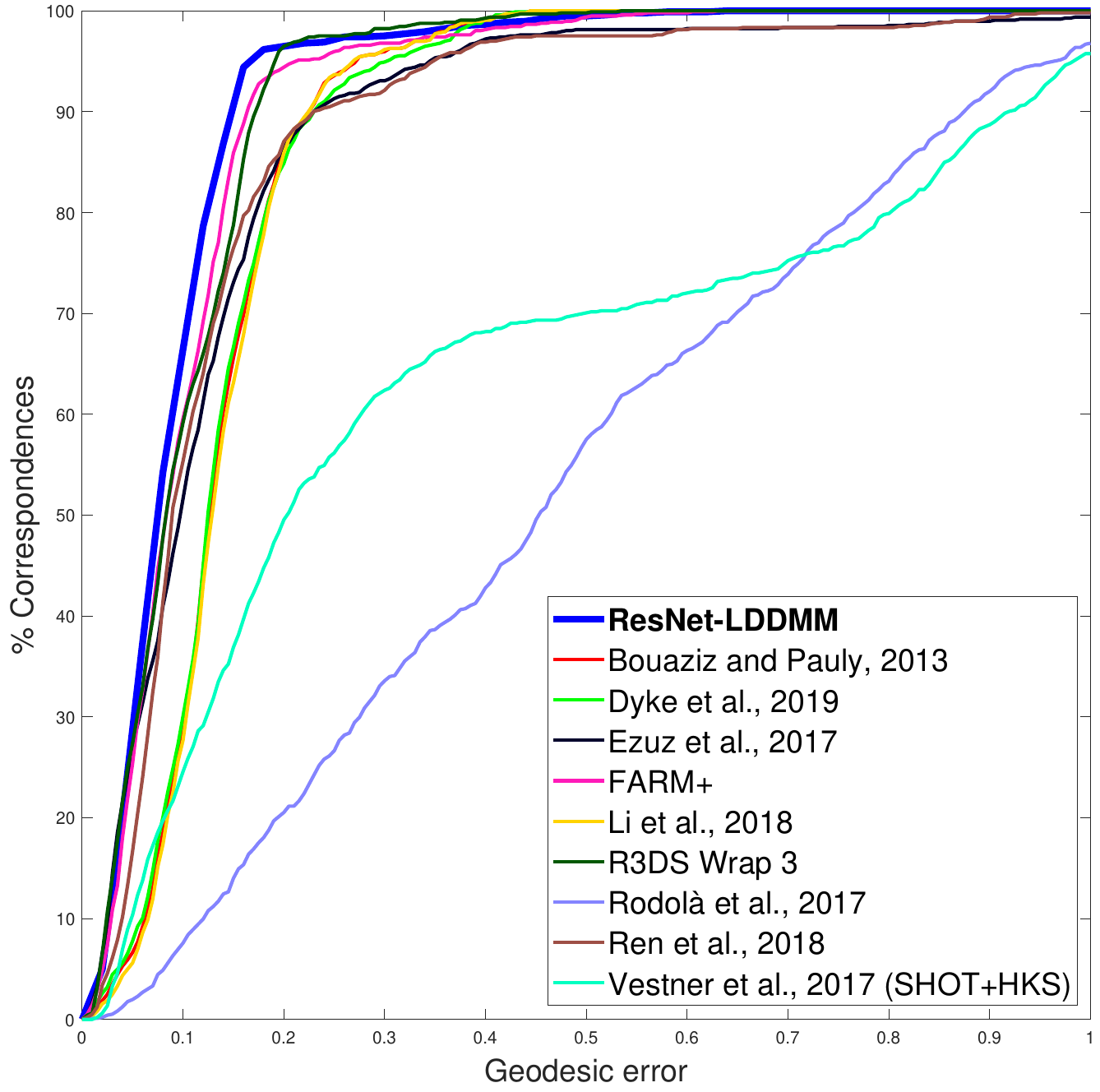}
    \includegraphics[width=0.19\linewidth]{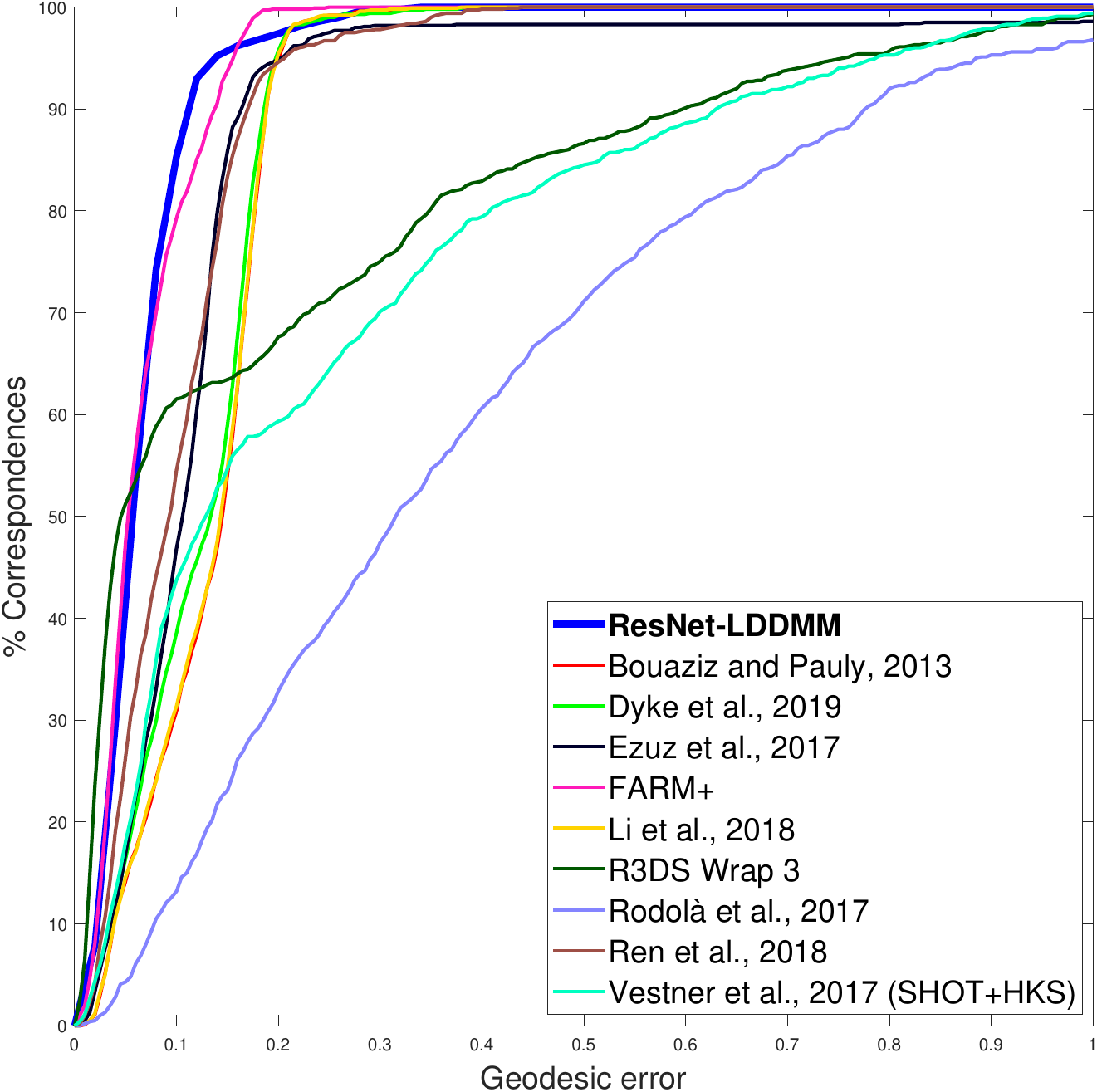}
    \includegraphics[width=0.19\linewidth]{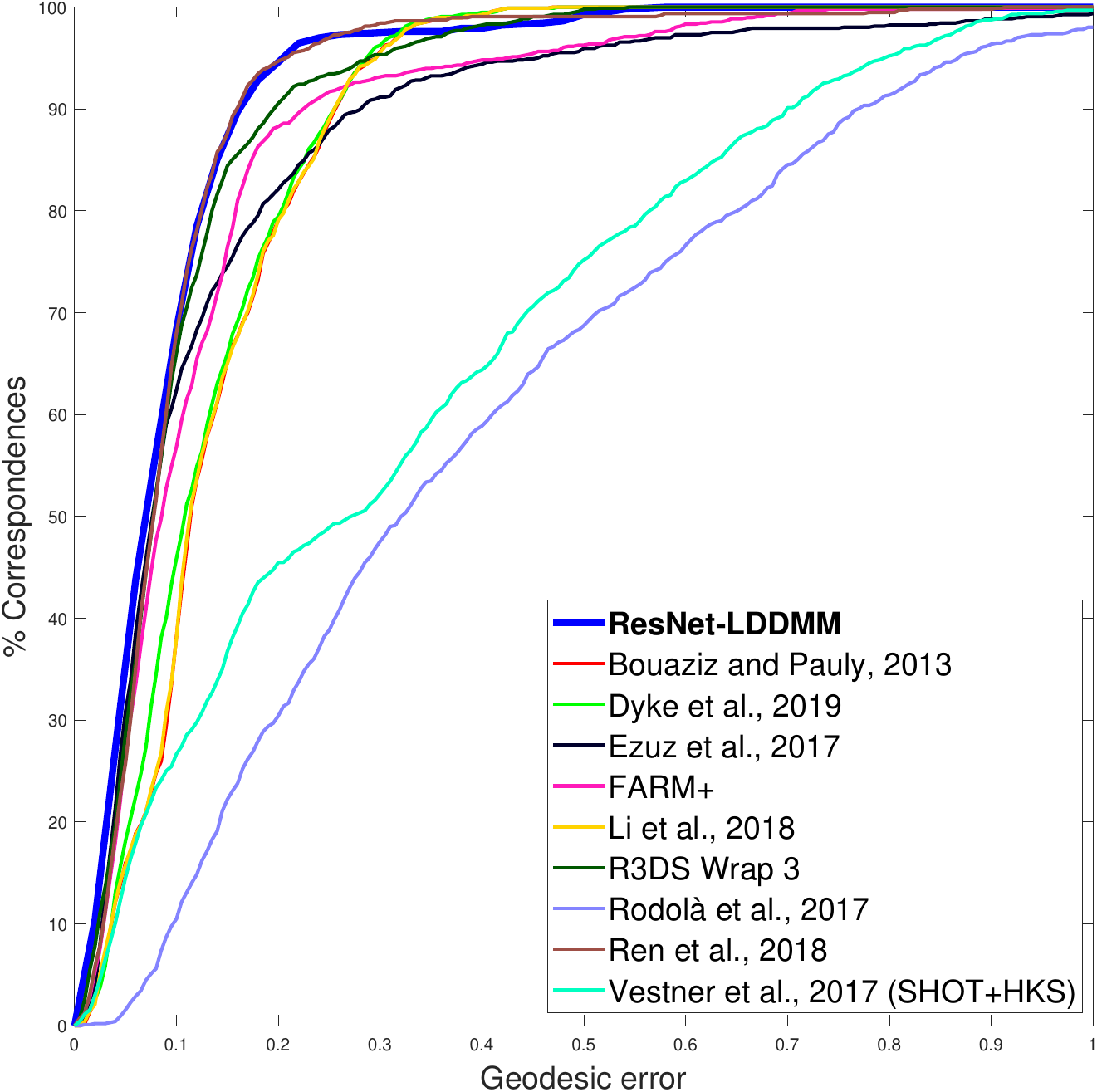}
    \includegraphics[width=0.19\linewidth]{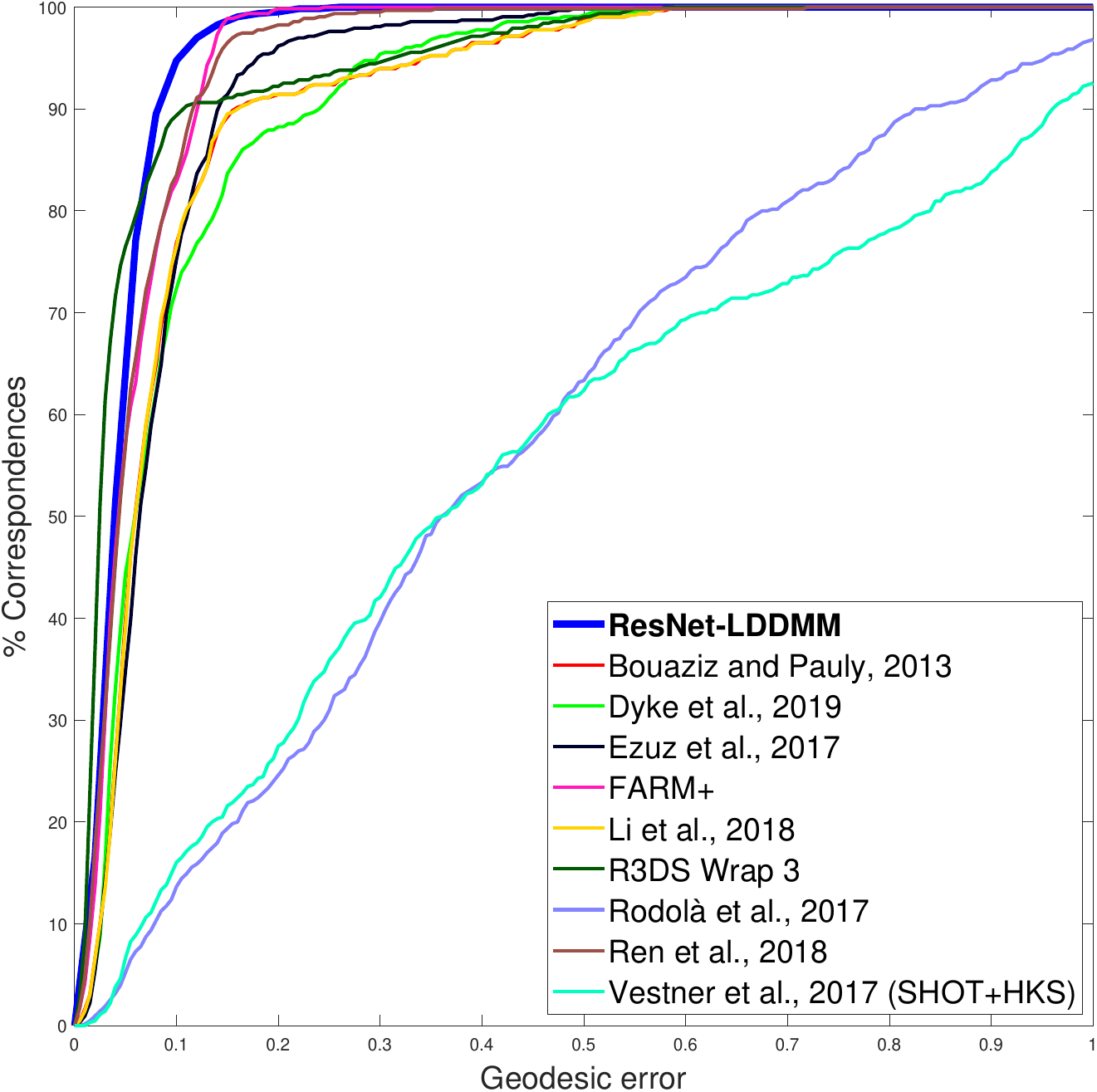}
    \includegraphics[width=0.19\linewidth]{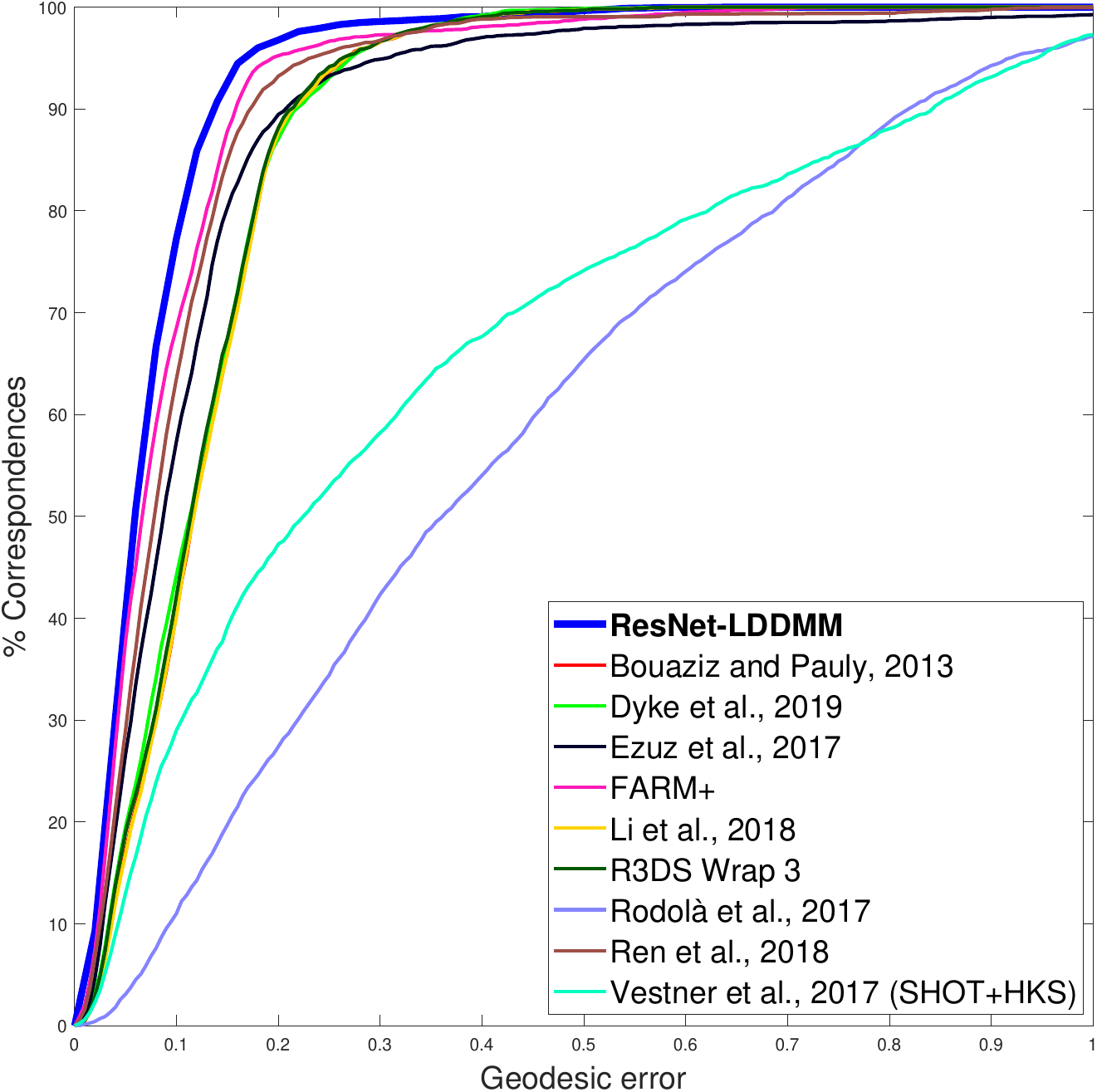}
    \vspace{-0.3cm}
\caption{Quantitative evaluation (Geodesic Error Curves) on SHREC'2020 and comparison with existing approaches including variants of the \textit{Functional Maps} framework, as reported in \cite{dyke2020shrec}. Considered poses/deformations, from left to right: twist, indent, inflate, stretch, and overall.}
   \label{Fig:SHREC2020}
\end{figure*}

The evaluation on test-set0 revealed another limitation of our ResNet-LDDMM. For example, when the hand's fingers are close to each others in the source shape and far in the target shape, ResNet-LDDMM fails to find the correct space partition in which affine transformations are computed. This dictates the need to incorporate the mesh connectivity in the current framework. That is only neighboring connected vertices could moves together to similar directions. In this experiment, prior to shape registration using our ResNet-LDDMM, we apply a Laplacian filtering on both source and target shapes. Empirically, we found that such pre-processing helps ResNet-LDDMM to find better registration solutions. We have used the simple nearest neighbor criterion to return the final correspondence after performing our ResNet-LDDMM algorithm.

\subsection{Experiments on SHREC'2020}
\label{Sec:SHREC2020}

We have also conducted experiments on the SHREC'2020 dataset (Task.1 \cite{dyke2020shrec}). Here, the dataset contains 11 partial scans and one full scan of a rabbit toy. Each scan represents a unique combination of pose (twist, indent, inflate or stretch) and internal materials (couscous, risotto, and chickpea). The challenge consists in registering each of the partial scans with the full (neutral pose) 3D scan. Unlike the Functional Maps framework which requires an adaptation as described in \cite{rodola2017partial}, the same implementation of ResNet-LDDMM was used to register partial scans to full scans. Unlike the previous SHREC'2019 contest, all participants have reported dense correspondence results. The graphs of Fig.\ref{Fig:SHREC2020} reports quantitative results and compares ResNet-LDDMM to existing approaches. Clearly on SHREC'2020, our approach outperforms all existing methods under all deformations and filling materials. In particular, it outperforms all variants of \textit{Functional Maps} with partial matching adaptation \cite{rodola2017partial}, with denoising by blurring the map \cite{ezuz2017deblurring} and the commercial product FARM+ dedicated to 3D human bodies, grounding on the Functional Maps framework \cite{ovsjanikov2012functional}. The key point is that \textit{Functional Maps} approaches works well on near-isometric deformations and suppose clean meshes (without noise or holes). ResNet-LDDMM tolerates such imperfections, works in partial matching and handles better non-isometric transformations (\eg stretching). To sum up, these results (graphs of Fig.\ref{Fig:SHREC2019} and Fig.\ref{Fig:SHREC2020}) show clearly that ResNet-LDDMM (1) tolerates noisy and missing data caused by self-occlusion or due to partial scans as it is works on point clouds and not meshes (more results and comparisons are reported in Supplementary Materials); (2) allows to go beyond near-isometric transformations and accounts for anisotropic deformations as twist, indent, inflate, and stretch. These evaluations reflects as well some limitations of ResNet-LDDMM which are (1) the restriction of ResNet-LDDMM to compute topology-preserving transformations (\ie diffeomorphic transformations) which do not allow accurate registration of meshes with different topologies; (2) the need to incorporate initial correspondences and points connectivity, when available, in order to better guide a correct division of the space into polytopes. Limitations and strengths to use diffeomorphisms (in LDDMM family of approaches) are discussed in the next paragraph.

\subsection{The Pros and Cons of using Diffeomorphisms}
\label{Sec:limitations}
While our approach differs from the usual LDDMM framework on a few points, it preserves a key aspect of these methods: it computes a full diffeomorphism of $\R^3$, and the template and target shapes themselves only appear in the loss function. This comes with both pros and cons, both also present in the original LDDMM methods, which we will now discuss.

\textbf{--} The main limitation of ResNet-LDDMM is that it will give poor result when trying to match shapes that are not topologically the same as embedded surfaces of $\R^3$. This is particularly visible when studying various poses of a hand or body: if two different parts of either shape touch in one case but not in the other (e.g., going from a closed hand to one pointing a finger), it is almost impossible to find an adequate diffeomorphism (examples reported in Fig.\ref{Fig:TopologyExamples}). The effect of this qualitative statement is shown quantitatively in Section \ref{Sec:SHREC2019} where we apply our algorithm to the SHREC'2019 database to mixed results, and visualized in Fig.\ref{Fig:TopologyExamples}. In such cases, methods that use an intrinsic approach, like Functional Maps, will perform better. 

\textbf{--} It should be noted that, depending on the goal, this ``drawback" can actually be a feature: a lot of applications of LDDMMs in computational anatomy come from its power to quantitatively tell how far one shape is from another, in addition to just matching them. This is reflected by how big the minimal value of the cost function is for various targets. In such cases, targets that are topologically different should indeed be considered very different, hence it being hard for them to be matched (see for example \cite{miller2007lddmm}).

\textbf{--} On the other hand, the well-known advantage of using diffeomorphisms is that when the shapes can be matched, it automatically ignores all transformations that do change the topology, preventing any unnecessary crossings in the shape. Moreover, because of the regularity of the transformation, it is generally very resistant to noise in the target shape. This has actually been used as a de-noising method for medical databases \cite{tward2013lddmm} (see also Fig.7 in the supplementary material). It also tolerates some missing data (see Fig.8 in the supplementary material for illustrations.

\textbf{--} Another perk of all LDDMM type methods, including ResNet-LDDMM, is that, since we obtain a diffeomorphism of all of $\R^3$, once trained, the networks can be applied to any point in the ambient space. This means that we can apply the trained transformation to a completely different parametrization of the shape. Consequently, it is possible to train the network on shapes with lowered resolutions and still get a good match for the original shapes.
\section{Conclusion}
\label{sec:Conclusion} 
Grounding on the elegant LDDMM framework, combined with the powerful Deep Residual Neural networks, we propose a joint geometric-neural network Riemannian-like framework for diffeomorphic registration of 3D shapes. Our registration schema is completely unsupervised, i.e. only source and target shapes are needed. A two-level regularization process was introduced, first by the network's structure and induced high-dimensional functional spaces and second by minimizing the time-integrated kinetic energy of the path connecting the source to the target. This allows the deformed (source) shape to move along its orbit under the group action of diffeomorphisms while tolerating an error defined by a data attachment term to the Target. Our ResNet-LDDMM geometrically builds time-dependent vector fields by first finding the optimal partition of the space into polytopes, and then predict affine transformations on each of these polytopes. Several experimental illustrations involving several challenges illustrate the ability of our framework computing diffeomorphic registrations with interesting computational complexity. The proposed framework also opens the door to design more efficient solutions in computational anatomy, statistical shape analysis and medical image registration. Just like LDDMM, our approach comes with a proper metric for shape registration and comparison. Finally, it should be noted that the loss function can be modified to make the training robust to a change of resolution or parametrization. This problem will be the subject of our future investigation.

\appendix[Proof of $f(\cdot,\theta^l)$ is Lipshitz]
\label{appendix:A1}

Each $f(\cdot,\theta^l)$ is Lipshitz, and for every $x,y$ in $\real^3$, we get
\begin{equation}
\begin{split}
&\Vert f(x,\theta^l) -f(y,\theta^l)\Vert_2\\
=&\left\Vert w_3^lw_2^l((w_1^l x+b_1^l)^+-(w_1^l y+b_1^l)^+)\right\Vert_2\\
\leq&\Vert w_3^l\Vert \Vert w_2^l\Vert\left\Vert((w_1^l x+b_1^l,0)^+-(w_1^l y+b_1^l,0)^+)\right\Vert_2,
\end{split}
\end{equation}
where $\Vert .\Vert $ is the matrix operator norm. Then, since the ReLU function is 1-Lipshitz, that is, $\Vert a^+-b^+ \Vert_2 \leq \Vert a-b \Vert_2$ for every $a,b$, we get
\begin{equation} \label{Eq:lipshitz}
\begin{split}
&\Vert f(x,\theta^l) -f(y,\theta^l)\Vert_2\\
\leq&\Vert w_3^l\Vert \Vert w_2^l\Vert \Vert w_1^l x+b_1^l- w_1^l y+b_1^l \Vert_2\\
\leq&\Vert w_3^l\Vert \Vert w_2^l\Vert \Vert w_1^l\Vert \Vert x- y \Vert_2.
\end{split}
\end{equation}

Now Eq. (\ref{Eq:discreteDiffeo}) is clearly a Euler scheme at the times $t=\Delta^L$ for the flow of a differential equation of the form
\begin{equation} \label{Eq:contDiffeo}\begin{split}
\forall x\in\real^3,\ t\in [0,1],\\ 
\Phi^0(x)=x\\
\partial_t\Phi(t,x)&=w_3(t)w_2(t)(\sigma(w_1(t)x+b_1(t))+b_2(t)),
 \end{split}
\end{equation}
with each coefficient of $w_i$ and each $b_j$ bounded in $t$. If we denote $f(t,x)=w_3(t)w_2(t)(\sigma(w_1(t)x+b_1(t))+b_2(t))$ the time-dependent vector field whose integration yields $\Phi(t,x)$, we immediately see that for each $t$, $f:x\mapsto f(t,x)$ is Lipshitz with, for every $t$, $x$ and $y$,
\begin{equation}\label{eq:LipshitzVF}\begin{split}
&\Vert f(t,x)-f(t,y)\Vert_2 \\
\leq &\Vert w_3(t)\Vert \Vert w_2(t)\Vert\Vert w_1(t)\Vert\Vert x-y\Vert_2\\\leq& \max_{t\in [0,1]} \Vert w_3(t)\Vert \Vert w_2(t)\Vert\Vert w_1(t)\Vert\Vert x-y\Vert_2,
\end{split}
\end{equation}
where $\Vert \cdot\Vert$ denotes the operator norm for linear operations, with the same reasoning as in Equation \eqref{Eq:lipshitz}. The Cauchy-Lipshitz theorem ensures that $\Phi$ exists for every $(t,x)$ and is (essentially) a diffeomorphism, or more precisely a bilipshitz transformation. In other words, ResNet-LDDMM does generate an approximation $\Phi^L$ of (essentially) a diffeomorphism $\Phi(1,\cdot)$ of $\real^3$, just like in LDDMM. As an additional remark, Gronwall's lemma shows that if we denote $C=\max_{t\in [0,1]} \Vert w_3(t)\Vert \Vert w_2(t)\Vert\Vert w_1(t)\Vert$, we have
\begin{equation}\begin{split}
    &\Vert\Phi(t,x)-\Phi(t,y)\Vert_2\leq \exp(tC)\Vert x-y\Vert_2,
    \\
   & t\in[0,1],\ x,y\in\real^3.
    \end{split}
\end{equation}
Going back to ResNet-LDDMM, a quick induction can be used to show that, denoting $C(\Theta)=\max_{l\in\{1,\dots,L\}}(\Vert w_3^l\Vert \Vert w_2^l\Vert\Vert w_1^l\Vert) $ with $\Theta=(\theta^1,\dots,\theta^L)$ an instance of ResNet-LDDMM, we have
\begin{equation}\begin{split}
    &\Vert f(x,\theta^l)-f(y,\theta^l)\Vert_2\leq C(\Theta)\Vert x-y\Vert_2,\ \text{and}\\
    &\Vert\Phi^l(x)-\Phi^l(y)\Vert_2\leq \exp(l\Delta^LC(\Theta))\Vert x-y\Vert_2,\\ 
    &l\in\{1,\dots,L\},\ x,y\in\real^3.
    \end{split}
\end{equation}
The regularizer term in the cost function (Eq. (\ref{Eq:ResNet-LDDMM})) then ensures that during the optimization, the constant $C(\Theta)$ does not go to infinity.

\section*{Acknowledgment}
The authors would like to thank the anonymous reviewers for their valuable comments and helpful suggestions. They are also thankful to \textit{Hao Hang}, Ph.D. student at NYU Abu Dhabi, for his valuable help conducting the evaluations on SHREC'2019 and SHREC'2020 datasets and \textit{Dr. Roberto Dyke} for providing the evaluation codes.


%





\ifCLASSOPTIONcaptionsoff
  \newpage
\fi



%

\bibliographystyle{ieeetr}
\bibliography{bare_jrnl_compsoc}

\end{document}